%% file: main.tex
\documentclass[acmsmall,nonacm,screen,11pt]{acmart}

\setcopyright{none}
\acmYear{2026}

\usepackage{brand}
\usepackage{header}
\usepackage{docspec}

\usepackage{pifont}
\newcommand{\cmark}{\ding{51}}
\newcommand{\xmark}{\ding{55}}
\usepackage{nicefrac}
\usepackage{subcaption}
\usepackage{makecell}
\usepackage{colortbl}
\usepackage{tabularx}
\usepackage{multirow}
\usepackage{enumitem}
\usepackage{placeins}
\usepackage{tikz}
\usetikzlibrary{positioning, calc, backgrounds, fit, arrows.meta}
\usepackage{pgfplots}
\usepackage{wrapfig}
\usepackage{needspace}
\pgfplotsset{compat=1.18}
\usepackage{booktabs}
\usepackage{longtable}

\usepackage{xspace}
\input{math_commands}

\input{abbreviations}
\input{non_english_phrases}

\newcommand{\methodname}{TerraTransfer++}

\begin{document}

\title[\methodname: Demonstration-Free End-to-End Planner]{\methodname: Demonstration-Free End-to-End Planner}

%% Structured authors (PDF metadata; rendering is the compact block below).
\author{Zikang Xiong}
\affiliation{\institution{Applied Intuition}\city{Sunnyvale}\state{CA}\country{USA}}
\author{Weixin Li}
\affiliation{\institution{Applied Intuition}\city{Sunnyvale}\state{CA}\country{USA}}
\author{Zhouchonghao Wu}
\affiliation{\institution{Applied Intuition}\city{Sunnyvale}\state{CA}\country{USA}}
\author{Akshay Rangesh}
\affiliation{\institution{Applied Intuition}\city{Sunnyvale}\state{CA}\country{USA}}
\author{Saarth Bonde}
\affiliation{\institution{Applied Intuition}\city{Sunnyvale}\state{CA}\country{USA}}
\author{Grantland Hall}
\affiliation{\institution{Applied Intuition}\city{Sunnyvale}\state{CA}\country{USA}}
\author{Chen Tang}
\affiliation{\institution{University of California, Los Angeles}\city{Los Angeles}\state{CA}\country{USA}}
\author{Yihan Hu}
\affiliation{\institution{Applied Intuition}\city{Sunnyvale}\state{CA}\country{USA}}
\author{Wei Zhan}
\affiliation{\institution{Applied Intuition}\city{Sunnyvale}\state{CA}\country{USA}}
\affiliation{\institution{University of California, Berkeley}\city{Berkeley}\state{CA}\country{USA}}

%% Compact rendering of the author block.
\compactauthors{%
  {\raggedright
   {\rmfamily\bfseries
    Zikang Xiong\textsuperscript{1}, \;
    Weixin Li\textsuperscript{1}, \;
    Zhouchonghao Wu\textsuperscript{1}, \;
    Akshay Rangesh\textsuperscript{1},\\[0.25em]
    Saarth Bonde\textsuperscript{1}, \;
    Grantland Hall\textsuperscript{1}, \;
    Chen Tang\textsuperscript{2}, \;
    Yihan Hu\textsuperscript{1}, \;
    Wei Zhan\textsuperscript{1,3,\dag}\par}
   \smallskip
   {\small\rmfamily\color{AppliedGray}%
    \textsuperscript{1}Applied Intuition \quad
    \textsuperscript{2}University of California, Los Angeles \quad
    \textsuperscript{3}University of California, Berkeley\par}
   \smallskip
   {\footnotesize\rmfamily\color{AppliedGray}%
    \textsuperscript{\dag}Corresponding author: wei.zhan@applied.co\par}
   \smallskip
   {\footnotesize\rmfamily\color{AppliedGray}%
    Project website: \url{https://terra-applied.github.io/TerraTransferPlusPlus/}\par}
   \par}
}

\begin{abstract}
End-to-end autonomous driving has achieved state-of-the-art performance on benchmarks and real-world deployments.
Its standard training recipe, however, is expensive across all existing paradigms: imitation learning requires fleet-scale driving logs whose behavior distribution only thinly covers the safety-critical tail;
closed-loop reinforcement learning in a photorealistic simulator, while offering diverse behaviors, is bottlenecked by sample-efficiency, per-step cost of rendering, and a forward pass through a large vision backbone.
Self-play in vectorized simulators changes the economics: at millions of rollout steps per second, one can obtain a state distribution with natural diversity in collisions, near-misses, and recoveries that no driving log contains, and a highly reactive policy emerges from a basic driving-quality reward.
Our approach exploits this asymmetry by decoupling \emph{learning to drive} and \emph{learning to plan} from \emph{learning to see}.
We pretrain a single policy by self-play, distill its rollouts into a flow-matching trajectory teacher that also plans, and train lightweight adapters on a frozen pretrained vision backbone to reproduce the latent space of either teacher through the action Kullback-Leibler divergence, a batch-relational low-rank structural loss, and, for the planner, a trajectory-matching loss.
Every target traces back to the self-play policy, so alignment never supervises against a logged trajectory: a paired dataset of (image, scene-state) frames suffices, with no need for the curated expert demonstrations that imitation pretraining is built on.
From this one self-play teacher we obtain two end-to-end students: a closed-loop policy that outperforms every published imitation-based end-to-end method on the photorealistic 3D Gaussian splatting closed-loop benchmark HUGSim, and a planner that reaches $59.4$ EPDMS on NAVSIM \texttt{navhard}, above every published method including DriveZero-Scale ($57.1$), and $95.5$ PDMS on \texttt{navtest}, without human demonstrations.
\end{abstract}
\maketitle

\clearpage
%% Table of contents: keep the entries clickable but unstyled (black, no
%% link color), so the contents page reads as plain text rather than a
%% block of brand blue.
\begingroup
\hypersetup{linkcolor=AppliedBlack}
\tableofcontents
\endgroup
\clearpage

\input{sections/1_intro}
\input{sections/3_method}
\input{sections/4_experiments}

\input{sections/2_related}
\input{sections/6_conclusion}

\bibliographystyle{ACM-Reference-Format}
\bibliography{references}

\newpage
\appendix
\input{sections/supplementary}

\end{document}

%% file: math_commands.tex
\usepackage{amsmath,bm} % amsfonts omitted: unicode-math (XeLaTeX) already provides \mathbb etc.

\def\eqref#1{equation~\ref{#1}}
\def\1{\bm{1}}

\def\rc{{\textnormal{c}}}

\def\vs{{\bm{s}}}

%% file: abbreviations.tex
\newcommand{\hugsim}{HUGSim\xspace}
\newcommand{\nuscenes}{nuScenes\xspace}
\newcommand{\nuplan}{nuPlan\xspace}
\newcommand{\pandaset}{PandaSet\xspace}
\newcommand{\kitti}{KITTI-360\xspace}
\newcommand{\waymo}{Waymo\xspace}

\newcommand{\hdscore}{HD-Score\xspace}

\newcommand{\wosac}{WOSAC\xspace}
\newcommand{\nhtsa}{NHTSA\xspace}

\newcommand{\terrazero}{TerraZero\xspace}

\newcommand{\uniad}{UniAD\xspace}
\newcommand{\vad}{VAD\xspace}
\newcommand{\ltf}{LTF\xspace}
\newcommand{\eco}{ECO\xspace}

\newcommand{\svd}{SVD\xspace}

\newcommand{\dac}{DAC\xspace}

\renewcommand{\rc}{\ensuremath{R_c}\xspace}

%% file: non_english_phrases.tex
\providecommand{\etc}{\emph{etc.}\xspace}

\renewcommand{\vs}{vs.\xspace} % math_commands.tex redefines \vs as a vector

\providecommand{\perse}{\emph{per se}\xspace}            % in itself / intrinsically
\providecommand{\tabularasa}{\emph{tabula rasa}\xspace}  % blank slate / from scratch

%% file: sections/1_intro.tex
% !TeX root = ../main.tex

\section{Introduction}
\label{sec:intro}

End-to-end autonomous driving maps raw sensor inputs~(\emph{e.g.}, camera
images) directly to control commands. Deep learning and large-scale driving
data have made this approach widely adopted and continuously improved across
benchmarks~\citep{bojarski2016end,codevilla2019cilrs,uniad,vad,ltf,navsim,hugsim}.
However, this progress relies on expensive standard pipelines that follow
either of the two routes.

The open-loop strategy draws its behavioral supervision from fleet logs:
imitation copies logged human
driving~\citep{codevilla2019cilrs,uniad,vad,ltf}, and open-loop reinforcement
learning (RL) optimizes against logged trajectories~\citep{karkus2025beyond}.
These logs provide millions of multi-camera
hours~\citep{caesar2020nuscenes,sun2020waymo,caesar2021nuplan}, yet
safety-critical events (collisions, near-misses, recoveries, \etc) remain
vanishingly rare in natural driving: at \nhtsa{} crash rates, a collision
occurs only once every ${\sim}16$k driving hours~\citep{nhtsa2023overview}, so
logging a single rare event can easily cost over \$1M in fleet operation
alone~\citep{kelkar2025sharedav}. Each sub-route inherits this scarcity in its
own way: imitation can only copy the actions the recorded drivers took, and
open-loop RL never sees how other vehicles would have responded to the
alternative actions it explores. Broadening interaction coverage with human
data therefore demands ever more collection and curation.

\begin{figure}[!t]
\centering
\pgfmathsetlengthmacro{\TeaserWidth}{\linewidth-0.8pt}
\pgfmathsetlengthmacro{\ConventionalWidth}{0.24*\TeaserWidth}
\pgfmathsetlengthmacro{\TeaserGap}{0.04*\TeaserWidth}
\pgfmathsetlengthmacro{\OursWidth}{0.72*\TeaserWidth}
\pgfmathsetlengthmacro{\OursImageWidth}{\OursWidth-16pt}
\begin{tikzpicture}[
    >=Stealth,
    font=\small\sffamily,
    conventional box/.style={draw=AppliedGray!65, rounded corners=3pt,
                line width=0.6pt, inner xsep=8pt, inner ysep=9pt,
                align=center, fill=AppliedGray!6},
    conventional arrow/.style={->, line width=0.7pt, draw=AppliedGray!80},
    outer frame/.style={rounded corners=7pt, line width=0.8pt},
    frame label/.style={font=\normalsize\sffamily\bfseries, fill=white,
                inner xsep=3pt, inner ysep=1pt, anchor=west},
    phase label/.style={font=\small\sffamily\bfseries, text=AppliedBlue,
                inner sep=0pt, anchor=north west},
  ]
  % The three horizontal allocations sum to the full usable line width.
  % Reserving one rule width keeps the rounded frame strokes within \linewidth.
  \coordinate (conventional-nw) at (0,0);
  \coordinate (conventional-ne) at ($(conventional-nw)+(\ConventionalWidth,0)$);
  \coordinate (ours-nw) at ($(conventional-ne)+(\TeaserGap,0)$);
  \coordinate (ours-ne) at ($(ours-nw)+(\OursWidth,0)$);

  % Right column: natural image aspect ratios determine the complete stack height.
  \node[frame label, text=AppliedBlue] (ours-label)
    at ($(ours-nw)+(7pt,0)$) {Ours};
  \node[phase label] (phase-one-label)
    at ($(ours-nw |- ours-label.south)+(8pt,-3pt)$)
    {Phase 1: learn to drive};
  \node[anchor=north west, inner sep=0pt] (phase-one)
    at ($(ours-nw |- phase-one-label.south)+(8pt,-4pt)$)
    {\includegraphics[width=\OursImageWidth]{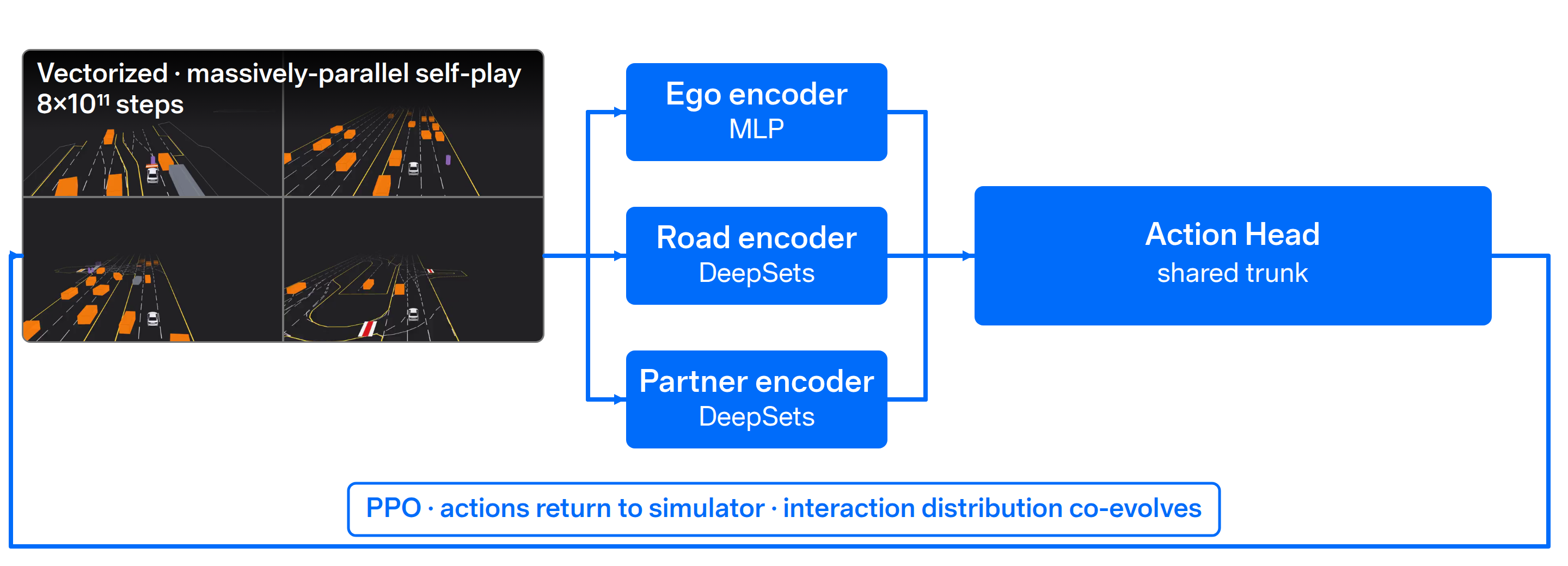}};
  \node[phase label] (phase-two-label)
    at ($(ours-nw |- phase-one.south)+(8pt,-7pt)$)
    {Phase 2: learn to plan};
  \node[anchor=north west, inner sep=0pt] (phase-two)
    at ($(ours-nw |- phase-two-label.south)+(8pt,-4pt)$)
    {\includegraphics[width=\OursImageWidth]{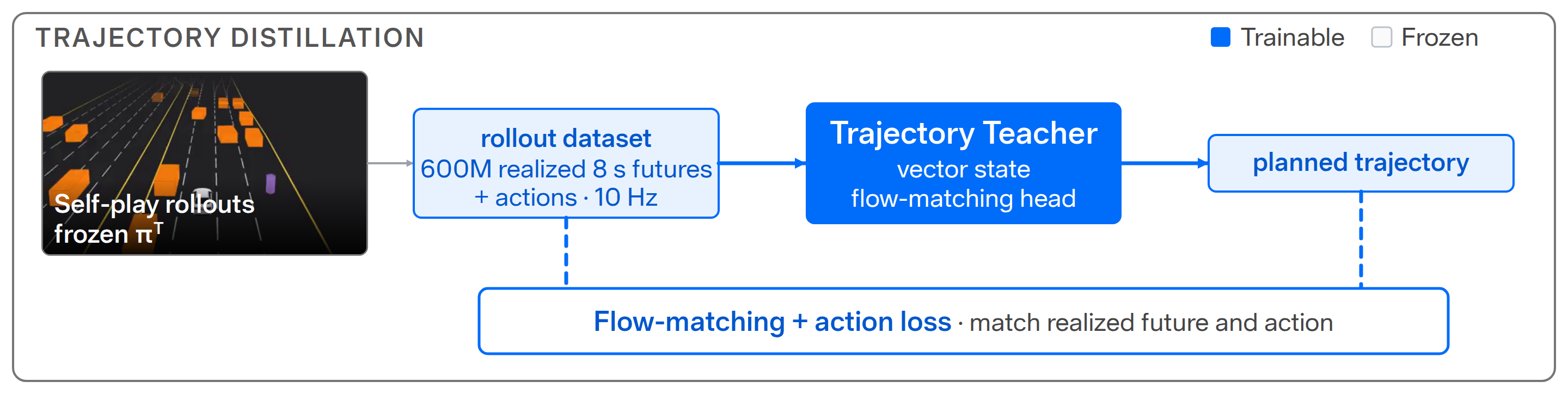}};
  \node[phase label] (phase-three-label)
    at ($(ours-nw |- phase-two.south)+(8pt,-7pt)$)
    {Phase 3: learn to see};
  \node[anchor=north west, inner sep=0pt] (phase-three)
    at ($(ours-nw |- phase-three-label.south)+(8pt,-4pt)$)
    {\includegraphics[width=\OursImageWidth]{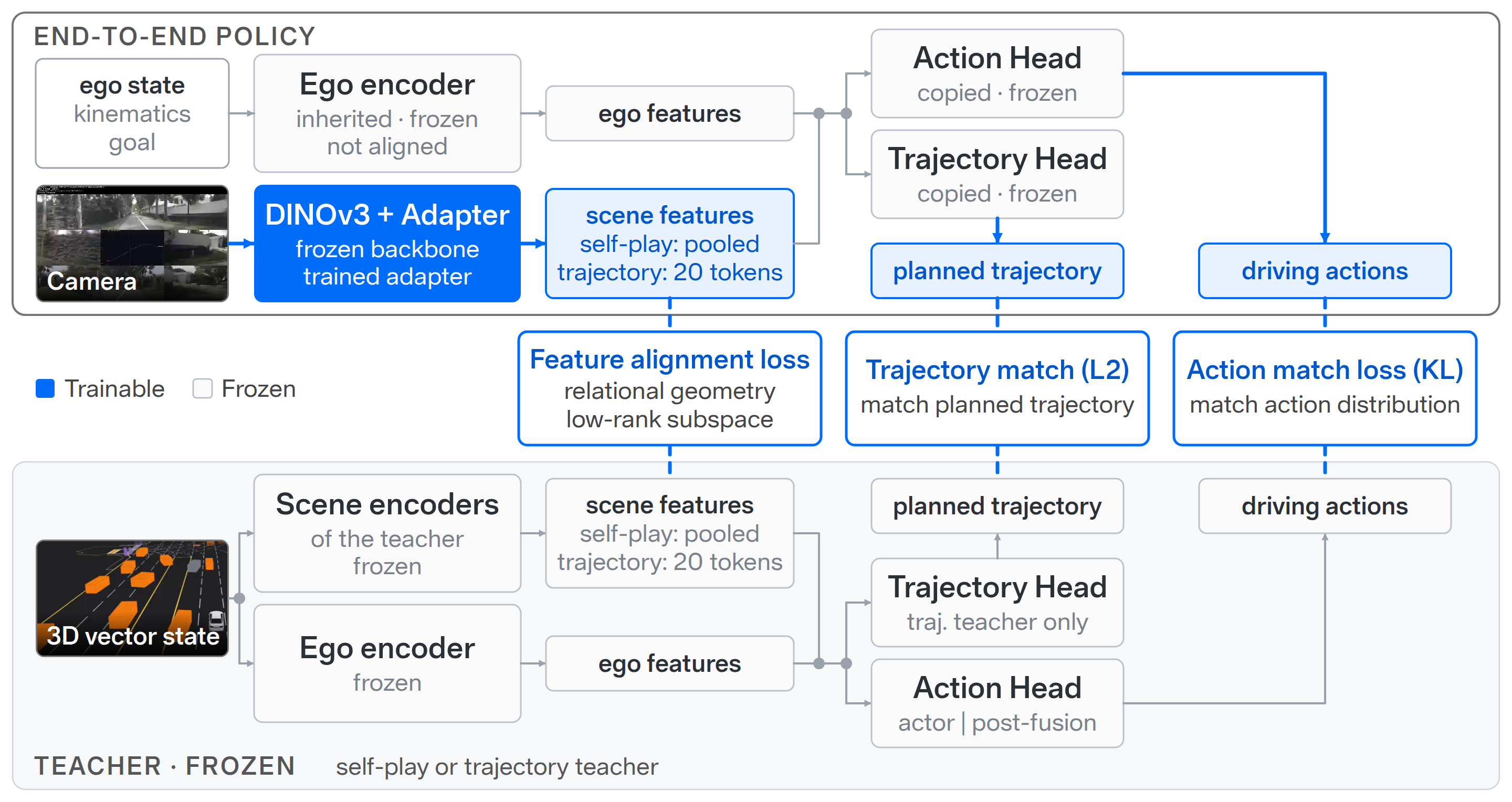}};
  \coordinate (frame-bottom) at ($(phase-three.south west)+(0,-8pt)$);

  % Distribute the four stages along the shared height instead of clustering them centrally.
  \coordinate (conventional-se) at (conventional-ne |- frame-bottom);
  \coordinate (conventional-top) at ($(conventional-nw)!0.5!(conventional-ne)$);
  \coordinate (conventional-bottom) at (conventional-top |- frame-bottom);
  \node[conventional box] (logs)
    at ($(conventional-top)!0.16!(conventional-bottom)$)
    {Fleet logs\\[-1pt]{\footnotesize\color{AppliedGray}human data}};
  \node[conventional box] (imitation)
    at ($(conventional-top)!0.39!(conventional-bottom)$)
    {Imitation\\[-1pt]pretraining};
  \node[conventional box] (finetuning)
    at ($(conventional-top)!0.62!(conventional-bottom)$)
    {Fine-tuning\\[-1pt]{\footnotesize\color{AppliedGray}SFT / RL}};
  \node[conventional box] (final-policy)
    at ($(conventional-top)!0.85!(conventional-bottom)$)
    {Final policy};
  \draw[conventional arrow] (logs.south) -- (imitation.north);
  \draw[conventional arrow] (imitation.south) -- (finetuning.north);
  \draw[conventional arrow] (finetuning.south) -- (final-policy.north);
  \node[frame label, text=AppliedGray] (conventional-label)
    at ($(conventional-nw)+(7pt,0)$) {Conventional};

  % Both frames use the same top baseline and the same image-derived bottom.
  \begin{scope}[on background layer]
    \draw[outer frame, draw=AppliedGray!55]
      (conventional-nw) rectangle (conventional-se);
    \draw[outer frame, draw=AppliedBlue!60]
      (ours-nw) rectangle (ours-ne |- frame-bottom);
  \end{scope}
\end{tikzpicture}
\caption{\textbf{Conventional training pipeline vs.\ our reinforcement-learning pretraining pipeline.} Conventional recipes imitate fleet logs and fine-tune on curated demonstrations, logged trajectories, or rendered simulation. They therefore need human data or rendering at every stage. \emph{Phase~1} (top right) trains a self-play policy $\pi_\theta$ with proximal policy optimization (PPO) in a multi-agent vectorized simulator. \emph{Phase~2} (middle right) freezes the policy as the self-play teacher and distills its rollouts into a trajectory teacher that also outputs a planned trajectory; both teachers read vector state. \emph{Phase~3} (bottom right) aligns an image-conditioned student to either frozen teacher: the self-play teacher yields the closed-loop student, and the trajectory teacher yields the planning student. The student keeps the teacher's ego encoder unchanged and is not aligned on it; only the teacher's scene features (road and partner, plus traffic for the trajectory teacher) enter the structural loss. An action loss matches the teacher's action distribution, while a structural loss matches its pairwise relationships among scenes; for the planning student, an L2 loss also matches the plan the trajectory teacher selects among $16$ samples. Alignment uses paired (image, scene-state) frames; it needs neither demonstrations nor online rendering during training. Blue modules are trainable; outlined modules are frozen. In the blue DINOv3 + Adapter block, only the adapter trains; the backbone stays frozen. The trajectory head and the L2 loss apply to the planning student only.}
\label{fig:teaser}
\label{fig:overview}
\end{figure}

The closed-loop framework instead trains inside a photorealistic,
sensor-rendering
simulator~\citep{dosovitskiy2017carla,chen2019lbc,zhang2021roach,wu2022tcp,ltf,karkus2025beyond},
where the policy experiences the consequences of its own actions. This
interaction, however, costs rendering and a heavy vision-backbone forward pass
at every step. Co-training multiple interacting agents multiplies these
per-step costs, so most image-based closed-loop work instead trains the ego
policy against scripted or log-replayed traffic, and the behavior
distribution stays close to the logged or hand-designed
content~\citep{dosovitskiy2017carla,chen2019lbc,zhang2021roach,wu2022tcp}.
These costs confine this route to fine-tuning a pretrained imitation policy,
as in Figure~\ref{fig:teaser}, rather than pretraining driving competence
from scratch.

Recently, multi-agent self-play has risen as a promising alternative to offer
the missing interaction: one shared policy controls every agent, so each
policy improvement reshapes the traffic every other agent faces, and the
interactive behavior co-evolves with the policy. Vectorized simulation makes
this co-training highly efficient, executing millions of steps per second on
commodity hardware~\citep{cusumano2025robust,chang2025spacer,seong2025grbo,guo2026correctionplanner}.
Training requires only a basic definition of driving quality as per-step
rewards: goal reaching, collision avoidance, comfort, and simple rule
compliance. These rewards specify general driving objectives rather than
separate rules for corner cases. The resulting interactions naturally produce
the collisions, recoveries, yielding, negotiation, and overtaking that logged
traffic seldom contains, all without expert trajectories or
behavior-specific rewards.

Despite its promise, however, self-play does not yield deployable end-to-end
driving policies \perse, as they observe vectorized scene state rather than
raw sensors. Deploying these policies would require a two-stage perception
stack that recovers their HD-map-style inputs from images. Such a stack
demands high-quality perception, is sensitive to perception noise, and forfeits
the co-optimization of a fully differentiable end-to-end model. As a result,
it has mostly remained confined to traffic simulation rather than an
end-to-end
solution~\citep{cusumano2025robust,chang2025spacer,seong2025grbo,guo2026correctionplanner,konstantinidis2025toward,ahmadi2026rlftsim}.

In this work, we unite the strengths of self-play and end-to-end driving in a
paradigm that learns driving policies without expert demonstrations. Self-play
learns driving competence without human demonstrations but consumes vector
state; end-to-end recipes consume raw sensors but depend on human
demonstrations, whose coverage of rare interactions is the very gap identified
above. We bridge this gap by decoupling \emph{learning to drive} and
\emph{learning to plan} from \emph{learning to see}, and we never use logged
demonstration trajectories as supervision. Figure~\ref{fig:teaser} contrasts its three stages with the
conventional workflow.

In the first stage, we train a policy $\pi_\theta$ \tabularasa{}
through multi-agent self-play in \terrazero~\citep{wu2026terrazero}, our
vectorized driving simulator, in the spirit of
GigaFlow~\citep{cusumano2025robust}. One parameter set controls every agent
in every scene. The policy learns from the long-tail distribution of its own
behavior and never imitates a human trajectory. The resulting policy is a
strong planner, scoring $94.19$ on \nuplan{} val14 and $70.87$ on the
interactive long-tail InterPlan suite, above every imitation-based planner we
are aware of, and $0.632$ human-similarity on \wosac{} 2023, on par with the
strongest learned sim agents.

In the second stage, the policy becomes the self-play teacher, and a
trajectory teacher learns to plan from its self-play rollouts. A closed-loop
policy emits one action per step, but planning interfaces and pseudo
closed-loop benchmarks score a multi-second trajectory. We therefore distill
the realized futures of these rollouts into the trajectory teacher through a
flow-matching head. Like the self-play teacher, it still reads vector state.

In the third stage, we freeze either teacher and align a vision student to
it: the self-play teacher yields the closed-loop student, and the trajectory
teacher yields the planning student. The student retains the teacher's head unchanged and
learns encoders that produce its scene features for road elements and
surrounding agents from images instead of vector states; these encoders pair
a pretrained vision backbone~\citep{simeoni2025dinov3} with learned adapters
supervised by paired (image, scene-state) frames of the same scene. The
teacher's per-state action distribution trains the student to choose the same
actions. A feature-level \emph{structural} loss preserves the teacher's
pairwise similarities among scene features, following the empirical finding
that the teacher's features are low-rank and redundant (\S\ref{sec:exp-rank}).
For the planning student, a trajectory loss also matches the plan the trajectory
teacher selects among $16$ samples, so the student learns the selection and needs
none at test time.
The closed-loop and planning students are two interfaces of one self-play
teacher, and neither sees a human trajectory.

Our framework preserves the naturalistic driving competence learned through
self-play. On photorealistic closed-loop benchmarks, the closed-loop student
attains an aggregate \hdscore{} of $0.548$, exceeding every published
end-to-end baseline on every \hugsim{} dataset split
(Table~\ref{tab:hugsim-all-datasets}). Likewise, on
NAVSIM, the planning student reaches $59.4$ EPDMS on the NAVSIMv2
\texttt{navhard} split, above every published method including
DriveZero-Scale ($57.1$), and $95.5$ PDMS on the NAVSIMv1 \texttt{navtest}
split, above every published sensor-based method
(Table~\ref{tab:navsim}). \methodname{} extends our earlier
TerraTransfer report~\citep{xiong2026terratransfer} with trajectory distillation, the second stage.

The cost of learning to drive moves upstream into self-play. This one-time
compute is efficient and reusable.
% 8 x 10^11 steps, 3m per step on average => 2.4 x 10^12 meters => 1.5B miles
The self-play policy is trained in 96 hours on 16 A100 GPUs, covering over 2.4B
kilometers of simulated driving without human demonstrations. It can then
supervise a range of vision frontends, alignment datasets, sensor stacks, and
downstream specializations.

During alignment, the vision encoder's only teacher signals are the frozen
head's per-state action distribution, the teacher's scene-feature relations,
and, for the planning student, the trajectory teacher's selected plan; all three
derive from the self-play teacher's behavior and are computed on a paired dataset of
(image, scene-state) frames. No expert behavior is required anywhere. Any
retained perception labels are policy-agnostic and reusable across frontends.
They are obtainable offline through auto-labeling, which separates this
requirement from the demonstration-collection burden of behavior cloning.
% 7129 shards x 256 frames per shard / 10hz / 3600s = 50.7 hours
In our case, the encoder is aligned over $1.83$M paired \nuplan{} frames plus
$300$K \hugsim{} domain-adaptation frames on 8 A100 GPUs in 10 hours. The
\nuplan{} paired frames are logged real camera images and the
domain-adaptation frames are rendered once offline, so training needs no
online rendering; rendering in the loop would be roughly $6{,}000\times$
slower than vector rollout ($175$k \vs $30$ fps across all cameras).

We make the following contributions:
\begin{itemize}
  \item \textbf{A novel demonstration-free paradigm for learning end-to-end
  driving policies.} A policy $\pi_\theta$ first learns to drive through
  multi-agent self-play in a vectorized simulator. Its per-state action
  distributions then supervise encoders that produce the policy's scene features
  from images, with no logged-trajectory supervision at any stage.
  \item \textbf{Trajectory distillation from self-play.} A flow-matching
  trajectory teacher learns the realized futures of the self-play teacher's
  rollouts, and the same alignment recipe plus a loss on the teacher's selected plan transfers it to a camera-only
  planner (\S\ref{sec:method:traj-distill} and \S\ref{sec:method:alignment}). The planner reaches $59.4$ EPDMS
  on NAVSIMv2 \texttt{navhard}, above every published method including
  DriveZero-Scale ($57.1$), and $95.5$ PDMS on NAVSIMv1 \texttt{navtest},
  above every published sensor-based method (Table~\ref{tab:navsim}).
  \item \textbf{A feature-level structural alignment mechanism.} We align the
  encoder directly with the teacher's pooled scene features instead of
  cascading reconstruction and detection. A batch-relational low-rank
  \emph{structural} loss matches the pairwise scene similarities induced by
  the teacher.
  \item \textbf{Demonstration-free closed-loop driving beyond published
  imitation-based end-to-end methods.} On photorealistic closed-loop scenarios
  in \hugsim, the camera-based closed-loop student, aligned to a teacher that
  never observed a human demonstration, outperforms every published
  imitation-based end-to-end method (Table~\ref{tab:hugsim-all-datasets}).
\end{itemize}

%% file: sections/3_method.tex
% !TeX root = ../main.tex

\section{Method}
\label{sec:method}

We first train a vector-state teacher through self-play, then distill its
rollouts into a trajectory teacher that also plans, and finally align a
vision student to either teacher with structural and action losses.

\subsection{TerraZero: Self-Play Vectorized Simulator}
\label{sec:method:terrazero}

Self-play runs in \terrazero~\citep{wu2026terrazero}, a vectorized driving
simulator that compiles \nuplan high-definition (HD) maps into static lane
topology, lane boundaries, and drivable-area geometry.

Training uses no \nuplan logged trajectories, logged agent initial states,
logged goals, human actions, or replayed traffic. Instead, controlled
vehicles spawn procedurally from lane centerlines through rejection sampling.
This procedure produces collision-free, drivable initial scenes.

Each controlled vehicle receives a procedural goal between $20$ and $100$\,m
ahead along the reachable lane graph. Reaching a goal triggers another sample
from the same topology. The simulator also places vulnerable road users and
static obstacles as scene context. The learned policy controls vehicles only.
Appendix~\ref{sec:supp-training-details} gives the complete environment and
training configuration.

At each $0.1$\,s step, a vehicle observes its ego kinematics and goal signal,
nearby road polylines and traffic controls, and neighboring scene entities. Each entity includes
relative pose, velocity, heading, size, type, and validity. These three input
groups become the ego, road, and partner streams used by the teacher below.
Appendix~\ref{sec:supp-observations} specifies the observation vectors.
Appendices~\ref{sec:supp-rewards}--\ref{sec:supp-dynamics-update} detail
rewards and dynamics.

\subsection{Self-Play Pretraining: The Teacher}
\label{sec:method:selfplay}

\paragraph{Setup.} We pretrain a driving policy $\pi_\theta$ through self-play
in \terrazero. A single set of parameters $\theta$ controls all agents in
every scene. We pool their transitions into one on-policy buffer. Each agent's
``surrounding traffic'' consists of other instances of the current policy. The
multi-agent interaction distribution therefore evolves with the policy itself.

\paragraph{Policy architecture.} At each step, an agent observes its own ego
state $o^{e}_t$. This state contains its kinematics, its goal signal, and the per-episode
reward-weight vector $w^{(n)}$ introduced below. The agent also observes a
variable-sized set of road elements
$\mathcal{O}^{r}_t=\{o^{r}_{t,i}\}_{i=1}^{M_r}$ and a variable-sized set of
partner agents $\mathcal{O}^{p}_t=\{o^{p}_{t,j}\}_{j=1}^{M_p}$. Three encoders
process these inputs in parallel:
\begin{equation}
\hat{f}^{e}_t = E_{\text{ego}}(o^{e}_t),\quad
\hat{f}^{r}_t = E_r(\mathcal{O}^{r}_t),\quad
\hat{f}^{p}_t = E_p(\mathcal{O}^{p}_t).
\label{eq:encoders}
\end{equation}
The ego encoder $E_{\text{ego}}$ is a multilayer perceptron (MLP), while
$E_r, E_p$ are set-processing DeepSets~\citep{zaheer2017deep} encoders. Each
DeepSets encoder max-pools its element embeddings. For the road encoder, the
pooling rule is $E_r(\mathcal{O}^{r}_t) = \mathrm{maxpool}_i\,\phi_r(o^{r}_{t,i}) + b_r$;
$E_p$ uses the same rule. The learned biases $b_r, b_p$ provide a well-defined
output when the corresponding set is empty. We concatenate the three features
and pass them through a shared MLP. An actor head then produces a categorical
distribution over a discrete set $\mathcal{A}$ of $12$ jerk actions (see
Appendix~\ref{sec:supp-vehicle-dynamics}):
\begin{equation}
\pi_\theta(\cdot \mid o_t) = \mathrm{softmax}\!\Big(W_a\cdot \mathrm{MLP}_{\text{shared}}\big([\hat{f}^{e}_t,\,\hat{f}^{r}_t,\,\hat{f}^{p}_t]\big)\Big).
\label{eq:actor}
\end{equation}

\paragraph{Reward.} Self-play supplies the experience; the reward defines
what counts as good behavior within it. We define the per-step reward
$R^{(n)}_t$ for agent $n$ as a weighted sum of $K$ driving-quality terms:
\begin{equation}
R^{(n)}_t \;=\; \sum\nolimits_{k=1}^{K} w^{(n)}_k \cdot r^{(k)}_t\!\big(o^{(n)}_t,\,a^{(n)}_t\big).
\label{eq:reward}
\end{equation}
The terms measure goal-reaching, collision, comfort, lane alignment, lane
centering, velocity, traffic-rule compliance, and related driving qualities
(see Appendix~\ref{sec:supp-rewards}).

Following the domain-randomization scheme of GigaFlow~\citep{cusumano2025robust},
we resample the weights $w^{(n)}_k\sim\mathcal{U}(w_k^{\min},w_k^{\max})$ for
each agent and episode. We include the sampled weights in the policy's ego
observation $o^{e}_t$. This conditioning trains a single $\pi_\theta$ to act
consistently across a continuum of driver preferences.

\paragraph{Objective.} We optimize the clipped proximal policy optimization
(PPO) surrogate~\citep{schulman2017proximal} with a value loss and entropy
bonus:
\begin{equation}
\mathcal{L}(\theta) = -\mathbb{E}_t\!\Big[\min\!\big(\rho_t\hat{A}_t,\;\mathrm{clip}(\rho_t,\,1{-}\epsilon,\,1{+}\epsilon)\hat{A}_t\big)\Big]
\;+\; c_v\,\mathcal{L}_V(\theta) \;-\; c_H\,\mathcal{H}\!\left[\pi_\theta(\cdot\mid o_t)\right].
\label{eq:ppo}
\end{equation}
Here, $\rho_t=\pi_\theta(a_t\mid o_t)/\pi_{\theta_{\text{old}}}(a_t\mid o_t)$ is
the policy probability ratio. The term $\hat{A}_t$ is the V-trace-corrected generalized advantage
estimation (GAE) advantage~\citep{schulman2015high}. The clipping threshold is
$\epsilon$. The coefficients $c_v$ and $c_H$ weight the value loss and entropy
bonus, respectively. After self-play, we freeze $\pi_\theta$ as the teacher
$\pi^T_\theta$.

\subsection{Trajectory Distillation}
\label{sec:method:traj-distill}

The frozen teacher $\pi^T_\theta$ emits one action per step but publishes no
planned trajectory, which planning interfaces and pseudo closed-loop
protocols require. We therefore give it a second interface: we distill its
rollouts into a trajectory teacher that also outputs a plan (Phase~2 of
Fig.~\ref{fig:overview}). Both teachers read privileged vector state;
\S\ref{sec:method:alignment} turns either into a vision policy.

\paragraph{Target construction.} For each frame of a self-play rollout, the
trajectory target is the realized future of $\pi^T_\theta$, not a logged human
trajectory. We record the next $80$ ego poses at the native $10$\,Hz control
rate ($8$\,s), each expressed in the current ego frame as forward offset,
lateral offset, and heading change, and standardize every timestep and
dimension with dataset statistics. A pose is masked out of the loss once the
future leaves the rollout horizon or the route switches. For NAVSIM, the
predicted plan is decimated to the $2$\,Hz, $4$\,s ($8$-pose) protocol
format~\citep{dauner2024navsim,cao2025navsimv2}. One hour of self-play
rollouts on $64$ GPUs yields $600$M such $8$\,s, $10$\,Hz trajectories for
trajectory distillation.

\paragraph{Trajectory teacher.} The trajectory teacher is a transformer
policy of $11.2$M parameters over the same vector observation, trained on the
actions and realized futures of $\pi^T_\theta$.
\emph{Entity encoders.} Unlike $\pi^T_\theta$, whose road stream also
carries the traffic controls, the trajectory teacher gives partner, road, and
traffic-control entities each their own encoder. An entity token is the sum of an MLP over its
features, a Fourier encoding of its relative position, and a type embedding
($d=128$). Masked self-attention runs over the tokens of each type ($2$ layers
for partners and roads, $1$ for traffic controls). Learned queries then
cross-attend to the tokens, and a linear projection with layer normalization
yields the per-type latent tokens: $8$ road latents
$z^{r}\in\mathbb{R}^{8\times64}$, $8$ partner latents
$z^{p}\in\mathbb{R}^{8\times64}$, and $4$ traffic latents
$z^{c}\in\mathbb{R}^{4\times64}$. A separate MLP encodes the ego state into
$e\in\mathbb{R}^{256}$. The heads see the scene only through these $20$ latent
tokens and $e$.
\emph{Fusion.} The latents are projected to $d=256$ and tagged with a type
embedding. Together with the ego token, they form $21$ tokens, which $3$
self-attention layers fuse into $F\in\mathbb{R}^{21\times256}$.
\emph{Action head.} An MLP on the fused ego token predicts the $12$ discrete
actions, trained with cross-entropy $\mathcal{L}_{\text{CE}}$ against the
action of $\pi^T_\theta$.
\emph{Trajectory head.} A flow-matching denoiser~\citep{lipman2023flow}
operates on the $80$ waypoint tokens. Each of its $4$ blocks applies
self-attention over waypoints, cross-attention to $F$, and an MLP. Every
block is modulated by adaptive layer normalization on the flow time $t$ and
$e$. Given a standardized target $x_1$, Gaussian noise $x_0$, and
$x_t=(1-t)x_0+t x_1$, the denoiser $v_\phi$ regresses the straight-line
velocity,
\begin{equation}
\mathcal{L}_{\text{FM}} = \mathbb{E}_{t,x_0}\Big[\textstyle\sum_{i} m_i\,\big\|v_\phi(x_t,t,F,e)_i-(x_{1,i}-x_{0,i})\big\|_2^2\Big]\big/\textstyle\sum_i m_i,
\label{eq:loss-fm}
\end{equation}
where $m_i$ masks invalid waypoints and $t$ is drawn from a logit-normal
distribution. The trajectory teacher minimizes
$\mathcal{L}_{\text{FM}}+\mathcal{L}_{\text{CE}}$. At inference, $10$ Euler
steps integrate $v_\phi$ from noise to a trajectory. We then replace its first
waypoint ($0.1$\,s) with the pose obtained by integrating the action head's
most likely action for one step through the jerk-actuated bicycle update of Appendix~\ref{sec:supp-dynamics-update},
so the plan starts with the action the policy executes. Every sampled plan
receives this replacement, including those compared by the trajectory loss
of \S\ref{sec:method:alignment}; it does not affect NAVSIM scores, whose
$2$\,Hz plans start at $0.5$\,s.

\subsection{Vision Alignment: The Student}
\label{sec:method:alignment}

\paragraph{Setup.} The alignment applies to either frozen teacher (Phase~3
of Fig.~\ref{fig:overview}). We describe it for $\pi^T_\theta$, which yields
the closed-loop student, and turn to the trajectory teacher at the end of
this section. The self-play teacher supplies
per-frame supervision. The student keeps the teacher's frozen policy head,
comprising the shared MLP and actor head. It also keeps the frozen ego encoder
$E_{\text{ego}}$. The student replaces only its own road and partner encoders
with image-conditioned counterparts $E^{\text{vis}}_r, E^{\text{vis}}_p$. Given an image $I_t$ paired with the
teacher's vector observation $o_t$, the student and teacher features are
\begin{equation}
\big(f^{r}_t,\,f^{p}_t\big) = \big(E^{\text{vis}}_r(I_t),\,E^{\text{vis}}_p(I_t)\big),\qquad
\big(\hat{f}^{r}_t,\,\hat{f}^{p}_t\big) = \big(E_r(\mathcal{O}^{r}_t),\,E_p(\mathcal{O}^{p}_t)\big).
\label{eq:align-feats}
\end{equation}
The frozen DeepSets encoders $E_r, E_p$ from Eq.~\ref{eq:encoders} produce the
teacher features $\hat{f}^{r}_t, \hat{f}^{p}_t$, exactly as during self-play.
The two vision encoders share a single frozen DINOv3~\citep{simeoni2025dinov3}
backbone and differ only in their adapters: $E^{\text{vis}}_r$ and
$E^{\text{vis}}_p$ apply separate linear adapters to the same backbone
features, producing the road and partner features $f^{r}_t, f^{p}_t$
(Fig.~\ref{fig:overview}). Under a fixed alignment recipe, this backbone gives
the strongest closed-loop score among the alternatives we ablate
(Appendix~\ref{sec:supp-backbone-ablation}).

The student receives camera images together with the ego vector, whose goal
signal is the endpoint of the logged ego trajectory, whereas the
teacher additionally receives vector road and partner state. A possible bridge
is a \emph{cascade}: a perception module recovers the teacher's vector inputs
(bounding boxes, lane polylines, and partner kinematics) from the image, and
the frozen teacher then consumes these recovered inputs. We avoid this
cascade for two reasons. First, it is not end-to-end differentiable: the
perception module trains on surrogate detection objectives, so no gradient
couples its errors to the driving quality that the frozen head serves, and the
perception stage makes any future end-to-end post-training of the policy
harder. Second, the teacher trains on clean simulator state and is sensitive
to input noise; a perception module that imperfectly recovers that state
degrades the head it feeds.

We supervise the student with two complementary objectives. The action loss
aligns policy distributions, but a softmax is coarse: different feature
configurations can yield the same action logits. The structural loss matches
feature relations in the informative low-rank subspace of
\S\ref{sec:exp-rank}. It preserves how the teacher arranges scenes relative
to one another, but its pairwise angles are invariant to rotations and
shifts, so it alone cannot ensure that the frozen head maps features to the
right actions. Together, the action loss anchors actions and the structural
loss shapes scene geometry.

\paragraph{Action alignment.} Following classical
distillation~\citep{hinton2015distilling}, we match the student's policy
distribution to the teacher's by forward KL divergence,
\begin{equation}
\mathcal{L}_{\text{act}} \;=\; \mathrm{KL}\!\big(\pi^T(\cdot\mid o_t)\;\|\;\pi^S(\cdot\mid I_t, o^e_t)\big).
\label{eq:loss-act}
\end{equation}
We match the full distribution rather than the sampled action: the
distribution carries the teacher's uncertainty across the action set, so
near-ambiguous scenes teach the student the same ambiguity instead of a single
arbitrary choice. This policy-level objective provides a global consistency
signal. Minimizing it encourages the student's policy distribution to approach
the teacher's, regardless of how the student represents road and partner
features.

\paragraph{Structural alignment.}
The structural loss matches scene geometry: scenes the teacher
treats as alike should remain alike for the student. This relational target
complements the action loss rather than adding a second coordinate-regression
objective.

The informative variation in the teacher's features is sharply low-rank, as
shown in \S\ref{sec:exp-rank}. Matching every coordinate would also force the
student to fit the structureless tail. An optimal low-rank denoiser discards
these low-variance directions as noise~\citep{gavish2014optimal}. We call the
span of the informative directions the \emph{teacher principal subspace} and
align features only within it.

For a batch of size $B$, we process each encoder $*\in\{r,p\}$ separately. We
first mean-center the student and teacher features, obtaining matrices
$\bar{F}_*, \bar{\hat{F}}_* \in \mathbb{R}^{B\times d}$. To extract the
teacher principal subspace, we compute a per-batch basis from the top-$k_*$
right singular vectors of the stop-gradient teacher matrix:
\begin{equation}
\bar{\hat{F}}_* \overset{\mathrm{SVD}}{=} U_*\Sigma_* V_*^\top,
\qquad
V_*^{(k_*)} = \big[\,v_{*,1},\dots,v_{*,k_*}\,\big]\in\mathbb{R}^{d\times k_*}.
\end{equation}
The basis $V_*^{(k_*)}$ spans the dominant axes of variation across the
current scenes. We then project both feature matrices onto the teacher
principal subspace. The projections are $Z_* = \bar{F}_* V_*^{(k_*)}$ and
$\hat{Z}_* = \bar{\hat{F}}_* V_*^{(k_*)}\in\mathbb{R}^{B\times k_*}$.

After unit-normalizing the projected features to remove per-scene magnitude
(both projections already have zero batch mean), we compare cosine similarities among
the $B$ scenes:
\begin{equation}
S_* = \tilde{Z}_*\,\tilde{Z}_*^\top, \quad
\hat{S}_* = \tilde{\hat{Z}}_*\,\tilde{\hat{Z}}_*^\top \in \mathbb{R}^{B\times B},
\qquad
\tilde{Z}_{*,i} = \frac{Z_{*,i}}{\lVert Z_{*,i}\rVert},
\qquad
\tilde{\hat{Z}}_{*,i} = \frac{\hat{Z}_{*,i}}{\lVert\hat{Z}_{*,i}\rVert}.
\end{equation}
Here, $\hat{S}_{*,ij}$ encodes the angle between scenes $i$ and $j$ in the
teacher principal subspace. This angle indicates which scenes the teacher
treats as alike. The student learns to replicate this geometry by minimizing
the squared Frobenius norm of the two matrices' difference:
\begin{equation}
\mathcal{L}^{(*)}_{\text{struct}} \;=\; \frac{1}{B^2}\big\lVert S_* - \hat{S}_* \big\rVert_F^2,
\qquad
\mathcal{L}_{\text{struct}} = \mathcal{L}^{(r)}_{\text{struct}} + \mathcal{L}^{(p)}_{\text{struct}}.
\end{equation}
We recompute the basis $V_*^{(k_*)}$ and target matrix $\hat{S}_*$ for each
batch under a stop-gradient. This procedure makes the student match pairwise
scene relationships only within the teacher principal subspace. It leaves
both the absolute coordinate frame and the orthogonal noise complement free.
We use $k_p=13$ and $k_r=9$, the $80\%$ cumulative-energy cutoffs of the
partner and road teacher spectra (\S\ref{sec:exp-rank}).

\paragraph{Total loss.} The alignment phase optimizes
$\mathcal{L}_{\text{align}} \;=\; \mathcal{L}_{\text{act}} \;+\;
\lambda\,\mathcal{L}_{\text{struct}}$. We set $\lambda=0.5$.

\paragraph{Trajectory student.} Replacing $\pi^T_\theta$ with the trajectory
teacher of \S\ref{sec:method:traj-distill} in the same recipe yields the
planning student, which we call the trajectory student. A frozen DINOv3 backbone with lightweight adapters
predicts the $20$ per-type latent tokens $(z^{r},z^{p},z^{c})$ from the image,
and the action and structural losses align them with the trajectory teacher's
latents: $\mathcal{L}_{\text{act}}$ is the KL divergence to the trajectory
teacher's action head, and $\mathcal{L}_{\text{struct}}$ is applied per type
to the flattened latents, with $\lambda=0.5$. A third loss, $\mathcal{L}_{\text{traj}}$, distills the teacher's selected plan.
On \texttt{navtrain}, the trajectory teacher samples $16$ plans per frame and keeps the
best under the constant-velocity selection of \S\ref{sec:exp-setup}, which reads
ground-truth agent states. The frozen trajectory head samples a plan from the
student's latents, and $\mathcal{L}_{\text{traj}}$ is the L2 distance between this plan
and the selected teacher plan over all $80$ waypoints. It is added to
$\mathcal{L}_{\text{align}}$ with weight $1$. The ego encoder, fusion layers, and both heads
stay frozen and consume the predicted latents; the ego embedding is computed
from the ego state and is never aligned. At inference, the student samples a
single plan and performs no selection.

%% file: sections/4_experiments.tex
% !TeX root = ../main.tex

\section{Experiments}
\label{sec:experiments}

We test one claim: vision policies distilled from a self-play teacher can
drive competitively in closed loop and plan competitively in pseudo closed
loop. No training stage touches a logged human trajectory. We first define
the setup and the two scorers (\S\ref{sec:exp-setup}) and present the
self-play teacher that anchors the pipeline (\S\ref{sec:exp-teacher});
the teacher sets the reference level. We then compare the student and the
teacher against end-to-end baselines on every \hugsim{} split and against
concurrent self-play-to-pixel systems on their $345$-scenario protocol
(\S\ref{sec:exp-headline}), and compare the
trajectory student against published planners on NAVSIM
(\S\ref{sec:exp-navsim}); the two students are the main results. Next, the insights and ablations of
\S\ref{sec:exp-ablations} show that, under matched data and compute,
self-play alignment outperforms imitation and a modular cascade: the
training paradigm sets the score, both alignment losses
are necessary and nearly lossless, and the low-rank structural target is
read from the teacher's spectrum rather than tuned. The appendix details the setup of the rank ablation
(Appendix~\ref{sec:supp-loss-ablation}) and ablates the vision backbone
(Appendix~\ref{sec:supp-backbone-ablation}).

\subsection{Setup}
\label{sec:exp-setup}
We evaluate on all $436$ scenarios of \hugsim~\citep{hugsim}, a photorealistic
3D Gaussian splatting closed-loop benchmark, spanning its four dataset splits
(\nuscenes $88$, \pandaset $127$, \kitti{} $113$, \waymo{} $108$), each
divided into Easy, Medium, Hard, and Extreme tiers. On \hugsim{} we report two scores
throughout: the closed-loop (CL) \hdscore, which scores the realized rollout,
and \hugsim's native scorer, which scores a planned trajectory. Our
single-step controller replans at each step, emitting one immediate jerk
action that we Euler-integrate into \hugsim's bicycle-model control signals
in causal closed form, without a second dynamics model
(Appendix~\ref{sec:supp-dynamics-update} specifies the conversion). The
closed-loop scorer executes only this first action per step and scores the
resulting rollout. The native scorer, in contrast, needs a full trajectory
that our controller does not publish, so we give it the \emph{decel}
stand-in plan: the realized first step followed by a comfort-bounded stop.
Appendix~\ref{sec:supp-cl-hdscore} defines both scorers in detail;
Appendix~\ref{sec:supp-self-play-policy-baseline} reports per-split,
per-tier subscores for the teacher and the student (baselines appear in
Table~\ref{tab:hugsim-all-datasets}).
Absolute score magnitudes are not directly comparable across dataset splits:
\pandaset ships no HD vector map, so its scorer evaluates \dac against a per-scene 3D
ground point cloud and \rc against the native centerline.

\paragraph{Alignment data.}
The paired dataset only teaches the vision frontend to reproduce the frozen
teacher's behavior. The headline run uses $\approx\!1.83$M paired (image,
scene-state) frames from \nuplan~\citep{caesar2021nuplan}: $7{,}129$ shards of
$256$ frames, or $\sim$51 hours at $10$\,Hz. At each frame, the target is the
teacher's action distribution on the reconstructed scene. The logged ego
trajectory is never used as a target; only its endpoint enters the student's
ego vector as the goal signal (\S\ref{sec:method:alignment}). Perception labels are policy-agnostic and can be
obtained offline, so this supervision is decoupled from expert behavior rather
than strictly weaker than demonstration collection. Beyond the \nuplan{}
corpus, we sample $300$K paired frames in total from the \hugsim{}
reconstructions of the four evaluation splits ($\sim$75K per split) for domain
adaptation, following the real-world-simulation
practice of SimScale~\citep{tian2026simscale}. These synthetic frames are rendered
from the evaluation scenes themselves but cover only $10\%$ of their total route length. The
paradigm ablation and the privileged probe in \S\ref{sec:exp-ablations} use
the \nuplan{} frames alone; the rank and backbone ablations use the headline
data, including the domain-adaptation frames. The trajectory student of
\S\ref{sec:exp-navsim} is aligned on NAVSIM \texttt{navtrain} only.

\paragraph{Teacher plans on NAVSIM.}
The NAVSIM protocol of \S\ref{sec:exp-navsim} scores a $4$\,s plan, which
we take from the trajectory teacher of \S\ref{sec:method:traj-distill}. In
each scene, its flow-matching head samples $16$ trajectories from $16$
different noise draws, which yields $16$ candidate plans. We then forecast the other agents over the same
horizon and keep the candidate that scores best on collision,
time-to-collision (TTC), and progress against that forecast. The default
teacher forecasts the agents by constant-velocity extrapolation; the
IDM-selection variant forecasts them with the intelligent driver
model~\citep{treiber2000idm}. We apply this procedure at both
\texttt{navhard} stages. Like the teacher itself, selection reads ground-truth
agent states, so both teacher rows of Table~\ref{tab:navsim} are privileged
references.

\subsection{Teacher Closed-Loop Results}
\label{sec:exp-teacher}

Before vision alignment, the self-play teacher outscores every published
imitation-based planner on the \nuplan-based planning benchmarks whose maps it
trains on. It
reaches $94.19$ on the \nuplan val14 suite~\citep{caesar2021nuplan}, compared
with $92.28$ for the best prior imitation planner. On the interactive long-tail
InterPlan suite~\citep{hallgarten2024interplan}, it reaches $70.87$, compared
with $63.66$. The teacher's exported rollouts also score $0.632$ on the WOSAC
2023 human-likeness benchmark~\citep{montali2023wosac}, within $0.09$ of the
$0.722$ logged-oracle reference and on par with the strongest learned sim
agents, without observing a human demonstration.

The teacher also transfers beyond its \nuplan training maps. It reaches an
aggregate closed-loop \hdscore of $0.647$ on the \hugsim \nuscenes split,
$0.710$ on \waymo, $0.554$ on \pandaset, and $0.363$ on \kitti{}, across $436$
scenarios ($0.562$ overall; Table~\ref{tab:teacher-hugsim-full}). For \kitti,
whose lanes and road edges we re-derive from the
reconstruction point cloud, the teacher reads the re-derived edges.
Appendix~\ref{sec:supp-self-play-policy-baseline} reports the per-tier
closed-loop breakdowns.

These transfer results position the teacher as a privileged reference. It drives
from vectorized scene state on HD maps: \nuscenes{} and \waymo{} ship their
own, and we generate them for \kitti{} and \pandaset{} as teacher inputs.

\subsection{Closed-Loop Comparison with End-to-End Baselines}
\label{sec:exp-headline}

Table~\ref{tab:hugsim-all-datasets} places closed-loop \hdscore in the left block
and native planned-trajectory \hdscore in the right block for every \hugsim
dataset split. All baseline entries score the same re-run rollouts from the
released checkpoints. The self-play teacher uses privileged vectorized scene
state, so we treat it as an upper-bound reference rather than a directly
comparable image-based method; the \emph{Ours} rows are the main
result.

\paragraph{Main results.}
Pooled over all $436$ scenarios and weighted by per-tier scenario counts, the
teacher reaches $0.562$ closed-loop and $0.542$ under the native scorer
(Table~\ref{tab:teacher-hugsim-full}). The aligned student retains roughly
$97.5\%$ of the teacher's closed-loop performance ($0.548$ closed-loop,
$0.528$ native) and exceeds every published baseline under both scorers. The
strongest baseline, \eco, trails the student by $0.120$ closed-loop and
$0.184$ native; the imitation-trained \ltf, \uniad, and \vad{} reach only
$0.266$, $0.244$, and $0.187$ closed-loop. The lead holds on every dataset
split, not just the aggregate: the student leads every baseline on the Easy,
Medium, and Hard tiers of every split and on the All aggregate of every split.
\kitti{} is the closest, $0.321$ vs.\ $0.310$ for \eco{} closed-loop. The
imitation-trained baselines trail far behind: none exceeds $0.41$ on any split
(best: \ltf{} at $0.404$ on \nuscenes), and all three fall below $0.11$
on \kitti{}. Under native scoring the ordering is unchanged, so the headline
ordering does not depend on the score variant.

\begin{table*}[tp]
\caption{\textbf{Closed-loop and native \hdscore{} across all four \hugsim
dataset splits.} The left block reports closed-loop \hdscore, and the right
block reports native planned-trajectory \hdscore; both blocks score the same
re-run rollouts. \uniad, \vad, and \ltf{} are re-run from their released
checkpoints on the current scenes and maps; \eco{} uses its released
checkpoint. Results are grouped by difficulty
tier for \nuscenes ($88$ scenarios), \pandaset ($127$), \kitti{} ($113$), and \waymo ($108$).
The student emits a single jerk action per step and is scored natively
with the decel stand-in plan. The
\textit{italicized} teacher row is the privileged reference and
is excluded from best-per-column counts.
Higher is better; \textbf{bold} marks the best comparison entry per
column within each split.}
\label{tab:hugsim-all-datasets}
\centering
\scriptsize
\renewcommand{\arraystretch}{1.0}
\setlength{\tabcolsep}{4pt}
\begin{tabular}{@{}lccccc|ccccc@{}}
\toprule
& \multicolumn{5}{c}{Closed-loop \hdscore} & \multicolumn{5}{c}{Native \hdscore} \\
\cmidrule(lr){2-6}\cmidrule(lr){7-11}
Method & Easy & Medium & Hard & Extreme & All & Easy & Medium & Hard & Extreme & All \\
\midrule
\multicolumn{11}{@{}l}{\textit{\nuscenes}} \\
\quad \uniad~\citep{uniad}                 & 0.423 & 0.231 & 0.286 & 0.122 & 0.259 & 0.412 & 0.234 & 0.290 & 0.129 & 0.260 \\
\quad \vad~\citep{vad}                     & 0.443 & 0.256 & 0.270 & 0.108 & 0.267 & 0.424 & 0.246 & 0.257 & 0.113 & 0.257 \\
\quad \ltf~\citep{ltf}                     & 0.708 & 0.440 & 0.326 & 0.111 & 0.404 & 0.702 & 0.440 & 0.331 & 0.124 & 0.407 \\
\quad \eco~\citep{eco}      & 0.764 & 0.416 & 0.405 & 0.255 & 0.452 & 0.846 & 0.431 & 0.403 & \textbf{0.263} & 0.476 \\
\midrule
\quad \textit{Teacher (ref.)}   & \textit{0.906} & \textit{0.684} & \textit{0.685} & \textit{0.279} & \textit{0.647} & \textit{0.875} & \textit{0.657} & \textit{0.653} & \textit{0.250} & \textit{0.618} \\
\quad \textbf{Ours}               & \textbf{0.901} & \textbf{0.682} & \textbf{0.641} & \textbf{0.278} & \textbf{0.636} & \textbf{0.884} & \textbf{0.650} & \textbf{0.620} & 0.251 & \textbf{0.610} \\
\midrule
\multicolumn{11}{@{}l}{\textit{\pandaset}} \\
\quad \uniad~\citep{uniad}                 & 0.577 & 0.380 & 0.235 & 0.125 & 0.295 & 0.576 & 0.384 & 0.242 & 0.135 & 0.301 \\
\quad \vad~\citep{vad}                     & 0.477 & 0.305 & 0.192 & 0.108 & 0.241 & 0.467 & 0.303 & 0.191 & 0.113 & 0.241 \\
\quad \ltf~\citep{ltf}                     & 0.882 & 0.376 & 0.208 & 0.085 & 0.311 & 0.877 & 0.386 & 0.219 & 0.097 & 0.323 \\
\quad \eco~\citep{eco}      & 0.884 & 0.639 & 0.174 & \textbf{0.247} & 0.446 & 0.876 & 0.470 & 0.044 & 0.156 & 0.326 \\
\midrule
\quad \textit{Teacher (ref.)}   & \textit{0.890} & \textit{0.883} & \textit{0.302} & \textit{0.224} & \textit{0.554} & \textit{0.876} & \textit{0.872} & \textit{0.275} & \textit{0.186} & \textit{0.532} \\
\quad \textbf{Ours}               & \textbf{0.892} & \textbf{0.871} & \textbf{0.353} & 0.234 & \textbf{0.565} & \textbf{0.882} & \textbf{0.855} & \textbf{0.324} & \textbf{0.189} & \textbf{0.539} \\
\midrule
\multicolumn{11}{@{}l}{\textit{\kitti}} \\
\quad \uniad~\citep{uniad}                 & 0.231 & 0.101 & 0.079 & 0.024 & 0.106 & 0.232 & 0.107 & 0.082 & 0.028 & 0.110 \\
\quad \vad~\citep{vad}                     & 0.084 & 0.068 & 0.075 & 0.026 & 0.063 & 0.079 & 0.063 & 0.071 & 0.029 & 0.061 \\
\quad \ltf~\citep{ltf}                     & 0.175 & 0.071 & 0.066 & 0.026 & 0.082 & 0.181 & 0.076 & 0.075 & \textbf{0.033} & 0.088 \\
\quad \eco~\citep{eco}                     & 0.530 & 0.432 & 0.126 & \textbf{0.108} & 0.310 & 0.501 & 0.282 & 0.019 & 0.008 & 0.205 \\
\midrule
\quad \textit{Teacher (ref.)}   & \textit{0.659} & \textit{0.497} & \textit{0.244} & \textit{0.019} & \textit{0.363} & \textit{0.650} & \textit{0.488} & \textit{0.205} & \textit{0.015} & \textit{0.348} \\
\quad \textbf{Ours}               & \textbf{0.550} & \textbf{0.457} & \textbf{0.219} & 0.017 & \textbf{0.321} & \textbf{0.535} & \textbf{0.450} & \textbf{0.195} & 0.014 & \textbf{0.309} \\
\midrule
\multicolumn{11}{@{}l}{\textit{\waymo}} \\
\quad \uniad~\citep{uniad}                 & 0.591 & 0.275 & 0.334 & 0.095 & 0.315 & 0.585 & 0.278 & 0.340 & 0.101 & 0.317 \\
\quad \vad~\citep{vad}                     & 0.237 & 0.197 & 0.208 & 0.109 & 0.189 & 0.226 & 0.189 & 0.199 & 0.114 & 0.183 \\
\quad \ltf~\citep{ltf}                     & 0.520 & 0.312 & 0.218 & 0.100 & 0.292 & 0.530 & 0.325 & 0.225 & 0.111 & 0.303 \\
\quad \eco~\citep{eco}                     & 0.801 & 0.581 & 0.339 & \textbf{0.254} & 0.510 & 0.757 & 0.463 & 0.213 & \textbf{0.124} & 0.403 \\
\midrule
\quad \textit{Teacher (ref.)}   & \textit{0.924} & \textit{0.854} & \textit{0.817} & \textit{0.115} & \textit{0.710} & \textit{0.912} & \textit{0.847} & \textit{0.777} & \textit{0.102} & \textit{0.694} \\
\quad \textbf{Ours}               & \textbf{0.941} & \textbf{0.836} & \textbf{0.758} & 0.116 & \textbf{0.695} & \textbf{0.936} & \textbf{0.827} & \textbf{0.717} & 0.100 & \textbf{0.679} \\
\bottomrule
\end{tabular}
\end{table*}

\begin{figure}[t]
\centering
\includegraphics[width=0.5\linewidth]{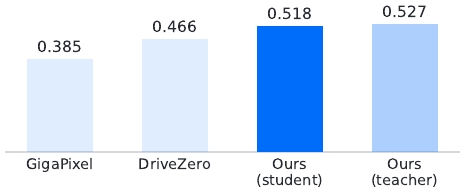}
\caption{\textbf{Native HD-Score on the shared $345$-scenario protocol.}
Our student exceeds both concurrent self-play-to-pixel systems; the
teacher upper-bounds the student. Our rows score the decel stand-in plan,
which satisfies the comfort bounds by construction; GigaPixel and DriveZero
are scored on their own plans.}
\label{fig:concurrent-comparison}
\end{figure}

\paragraph{Comparison with concurrent self-play-to-pixel systems.}
Two concurrent self-play-to-pixel systems report results on \hugsim{}:
GigaPixel~\citep{rowe2026gigapixel} trains in a dedicated bounding-box
renderer and reaches a native HD-Score of $0.385$;
DriveZero~\citep{he2026drivezero} regresses teacher rollouts collected on
real log-initialized worlds and reaches $0.466$. Both were evaluated on the
pre-challenge $345$-scenario protocol of \citet{kirby2026drivor}, which the
current $436$-scenario release later superseded; their checkpoints are not
public, so we re-evaluate our system on the same $345$ scenarios with
\hugsim's native scorer (Figure~\ref{fig:concurrent-comparison}). The
teacher reaches $0.527$ and the student $0.518$, exceeding both systems.
Our native scores use the decel stand-in plan, which satisfies the comfort
bounds by construction, whereas GigaPixel and DriveZero are scored on the
plans their policies output.
The training loops differ as well: GigaPixel keeps a renderer in its loop,
and DriveZero keeps per-scene teacher rollouts, whereas our alignment
consumes the frozen teacher's action distributions and scene-feature
relations on fixed paired frames: it needs no rendering in the loop and no
teacher rollouts during alignment (the $300$K domain-adaptation frames are
rendered once, offline).

\begin{figure}[t]
\centering
\includegraphics[width=\linewidth]{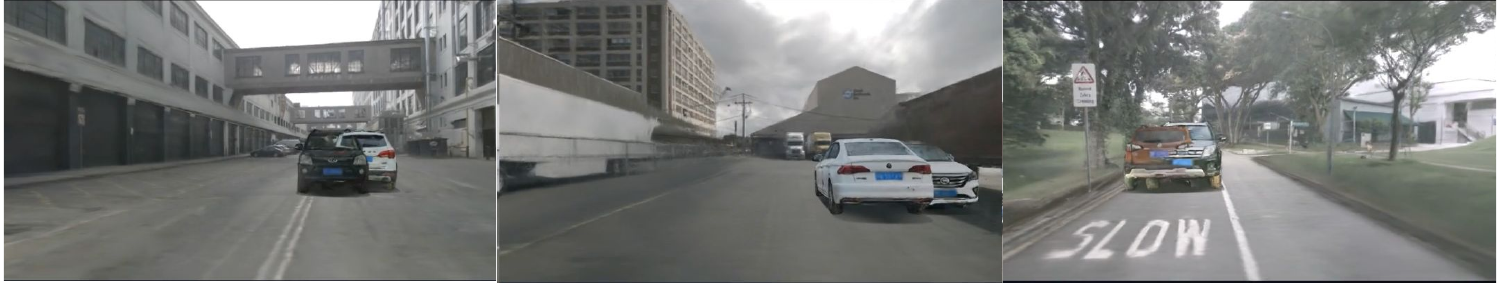}
\caption{\textbf{Visually implausible extreme-tier scenarios.} Some \hugsim{}
Extreme scenarios begin from implausible configurations: other views are
occluded, and vehicles appear to pass through one another. Configurations
like these trigger the student's conservative yielding, which sacrifices
route completion and pulls the Extreme scores of
Table~\ref{tab:hugsim-all-datasets} down for every method.}
\label{fig:supp-qualitative-extreme}
\end{figure}

\paragraph{Extreme-tier failure mode.}
The student's weakest tier is Extreme ($0.162$ closed-loop, $0.137$ native,
pooled over all splits), where even the teacher reference manages only
$0.158$ and $0.136$, and baselines retain an edge on parts of this tier:
\eco{} on \pandaset, \kitti, and \waymo{} under the closed-loop scorer and on
\nuscenes{} and \waymo{} under the native scorer, and the imitation-trained
baselines on \kitti{} under both. Its
adversarial scenarios are partly implausible
(Figure~\ref{fig:supp-qualitative-extreme}): the other agents collide with the
ego regardless of what the ego does, so a policy that just accelerates and
drives straight, never reacting, already scores $0.227$ closed-loop, above
every reactive policy. The student, like the teacher it is aligned to,
instead waits for the oncoming vehicle as a human driver would and
sacrifices route completion. The resulting low \rc reduces aggregate
\hdscore on this tier (Appendix~\ref{sec:supp-self-play-policy-baseline}).
The tier thus punishes exactly the reactive behavior we want.

\subsection{Pseudo Closed-Loop Planning on NAVSIM}
\label{sec:exp-navsim}

The closed-loop student publishes no plan, so \hugsim's native scorer sees
only the decel stand-in of \S\ref{sec:exp-setup}. We therefore evaluate the
trajectory student of \S\ref{sec:method:alignment} with the pseudo closed-loop NAVSIM protocol,
which scores a planned $4$\,s trajectory instead of a realized rollout. We
report \textbf{PDMS} on the NAVSIMv1 \texttt{navtest}
split~\citep{dauner2024navsim} and the two-stage \textbf{EPDMS} on the
NAVSIMv2 \texttt{navhard} split~\citep{cao2025navsimv2}.
Table~\ref{tab:navsim} groups the published baselines by input and
supervision; privileged planners that read ground-truth scene state, including our self-play
teacher, appear as italicized references.
Appendix~\ref{sec:supp-navsim-subitems} breaks both aggregates into their
subscores.

Table~\ref{tab:navsim} shows that the teacher's competence carries over to
the planning interface. The student samples one plan per scene with no
selection; selection enters only through its training target
(\S\ref{sec:method:alignment}), so the student amortizes the teacher's
best-of-$16$ choice. On \texttt{navhard}, the student reaches $59.4$ EPDMS
without human trajectory supervision, the highest score of any sensor-based
entry, $2.3$ points above DriveZero-Scale ($57.1$) and $2.7$ below the
privileged, selection-based trajectory teacher ($62.1$). On \texttt{navtest}, it reaches $95.5$
PDMS, above every sensor-based entry, including DriveZero-Scale ($95.3$), and
within $0.4$ points of the privileged best-of-$16$ trajectory teacher. The student's weak points, safety margins
and perturbed recovery, mirror its teacher's: its \texttt{navtest} TTC is
$92.2$ (teacher $93.2$, DriveZero-Scale $97.3$), and its Stage~2 DAC on
\texttt{navhard} is $90.2$ (teacher $90.4$, DriveZero-Scale $95.6$). IDM selection
(\S\ref{sec:exp-setup}) raises the teacher to $65.6$ EPDMS on
\texttt{navhard} and leaves \texttt{navtest} at $95.9$ PDMS.

\begin{table}[!t]
\centering
\caption{\textbf{Pseudo closed-loop planning on NAVSIM.} \texttt{navtest}
reports PDMS (NAVSIMv1); \texttt{navhard} reports the per-stage scores and
the two-stage EPDMS (NAVSIMv2).
Baseline rows are the published results compiled by
\citet{he2026drivezero}; -- marks a score that compilation does not list.
\emph{Demo} marks whether human trajectory demonstrations were used as
training supervision; \emph{Priv.} marks whether the planner reads
privileged ground-truth scene state (vectorized agents and map) at test time
rather than sensor input. \textit{Italicized} rows are privileged references,
not comparison entries. Bold marks the best score in each group. Our rows are
the trajectory teacher of \S\ref{sec:method:traj-distill} and the trajectory
student of \S\ref{sec:method:alignment}; neither uses human trajectory
supervision. Both teacher rows keep the best of $16$ sampled plans
(\S\ref{sec:exp-setup}); the student row scores a single sampled plan with no
selection, so it reads no ground-truth state; \emph{IDM selection} forecasts the other agents with
the intelligent driver model instead of constant-velocity extrapolation.}
\label{tab:navsim}
\footnotesize
\renewcommand{\arraystretch}{0.98}
\setlength{\tabcolsep}{6pt}
\begin{tabular}{@{}lcccccc@{}}
\toprule
 & & & \texttt{navtest} & \multicolumn{3}{c}{\texttt{navhard}} \\
\cmidrule(lr){4-4}\cmidrule(l){5-7}
Method & Demo & Priv. & PDMS$\uparrow$ & Stage 1$\uparrow$ & Stage 2$\uparrow$ & EPDMS$\uparrow$ \\
\midrule
\multicolumn{7}{@{}l}{\textit{Privileged reference (ground-truth scene state)}} \\
\quad \textit{Human driver} & -- & -- & \textit{94.8} & -- & -- & -- \\
\quad \textit{PDM-Closed~\citep{dauner2023pdm}} & \cmark & \cmark & \textit{89.1} & -- & -- & \textit{56.6} \\
\quad \textit{DriveRL~\citep{he2026drivezero}} & \xmark & \cmark & \textit{95.8} & \textit{84.1} & \textit{65.4} & \textit{55.6} \\
\quad \textit{Trajectory teacher (ours)} & \xmark & \cmark & \textit{\textbf{95.9}} & \textit{92.2} & \textit{67.5} & \textit{62.1} \\
\quad \textit{Trajectory teacher (ours, IDM selection)} & \xmark & \cmark & \textit{\textbf{95.9}} & \textit{\textbf{94.6}} & \textit{\textbf{69.5}} & \textit{\textbf{65.6}} \\
\midrule
\multicolumn{7}{@{}l}{\textit{Camera \& LiDAR input, human demonstrations}} \\
\quad TransFuser~\citep{ltf} & \cmark & \xmark & 84.0 & -- & -- & -- \\
\quad DRAMA~\citep{drama2025} & \cmark & \xmark & 85.5 & -- & -- & -- \\
\quad VADv2~\citep{vadv2} & \cmark & \xmark & 86.2 & -- & -- & -- \\
\quad DiffusionDrive~\citep{diffusiondrive} & \cmark & \xmark & 88.1 & -- & -- & -- \\
\quad Hydra-MDP++~\citep{hydramdp} & \cmark & \xmark & 91.0 & -- & -- & -- \\
\quad Centaur~\citep{centaur2025} & \cmark & \xmark & 92.6 & -- & -- & -- \\
\quad DriveSuprim~\citep{drivesuprim} & \cmark & \xmark & \textbf{93.5} & -- & -- & -- \\
\midrule
\multicolumn{7}{@{}l}{\textit{Camera input, human demonstrations}} \\
\quad \uniad~\citep{uniad} & \cmark & \xmark & 83.4 & -- & -- & -- \\
\quad PARA-Drive~\citep{paradrive} & \cmark & \xmark & 84.0 & -- & -- & -- \\
\quad Epona~\citep{epona} & \cmark & \xmark & 86.2 & -- & -- & -- \\
\quad OneVL~\citep{onevl} & \cmark & \xmark & 88.4 & -- & -- & -- \\
\quad DriveLaW~\citep{drivelaw} & \cmark & \xmark & 89.1 & -- & -- & -- \\
\quad AutoVLA~\citep{autovla} & \cmark & \xmark & 89.1 & -- & -- & -- \\
\quad DriveVLA-W0~\citep{drivevlaw0} & \cmark & \xmark & 90.2 & -- & -- & -- \\
\quad Qwen-Drive-1.0~\citep{qwendrive} & \cmark & \xmark & 90.7 & -- & -- & -- \\
\quad ReCogDrive~\citep{recogdrive} & \cmark & \xmark & 90.8 & -- & -- & -- \\
\quad R2SE~\citep{r2se} & \cmark & \xmark & 91.6 & -- & -- & -- \\
\quad iPad~\citep{ipad2026} & \cmark & \xmark & 91.7 & -- & -- & -- \\
\quad DriveFine~\citep{drivefine} & \cmark & \xmark & 91.8 & -- & -- & -- \\
\quad \ltf~\citep{ltf,dauner2024navsim} & \cmark & \xmark & -- & -- & -- & 25.1 \\
\quad NavFormer~\citep{navformer} & \cmark & \xmark & -- & -- & -- & 34.1 \\
\quad RAP~\citep{feng2025rap} & \cmark & \xmark & -- & -- & -- & 39.6 \\
\quad GuideFlow~\citep{guideflow} & \cmark & \xmark & -- & -- & -- & 51.5 \\
\quad SimScale~\citep{tian2026simscale} & \cmark & \xmark & -- & -- & -- & 53.2 \\
\quad DrivoR-Scale~\citep{kirby2026drivor} & \cmark & \xmark & \textbf{94.6} & -- & -- & \textbf{54.6} \\
\midrule
\multicolumn{7}{@{}l}{\textit{Camera input, no human demonstrations}} \\
\quad ZTRS~\citep{ztrs} & \xmark & \xmark & -- & -- & -- & 48.1 \\
\quad GigaPixel~\citep{rowe2026gigapixel} & \xmark & \xmark & -- & 77.8 & 63.5 & 50.1 \\
\quad DriveZero~\citep{he2026drivezero} & \xmark & \xmark & 94.8 & 84.1 & 60.8 & 51.5 \\
\quad DriveZero-Scale~\citep{he2026drivezero} & \xmark & \xmark & 95.3 & 82.3 & \textbf{69.1} & 57.1 \\
\quad \textbf{Ours} & \xmark & \xmark & \textbf{95.5} & \textbf{89.1} & 66.5 & \textbf{59.4} \\
\bottomrule
\end{tabular}
\end{table}

\subsection{Insights and Ablations}
\label{sec:exp-ablations}

Together, \S\ref{sec:exp-headline} and \S\ref{sec:exp-navsim} show that one
alignment recipe carries the self-play teacher's competence into both
students. The insights and ablations below study that recipe on the
closed-loop student of \S\ref{sec:method:alignment}.

\begin{figure}[t]
\newsavebox{\rankplotbox}
\centering
\captionsetup{font=small,position=below,skip=6pt,belowskip=0pt}
\savebox{\rankplotbox}{\includegraphics[width=0.36\linewidth]{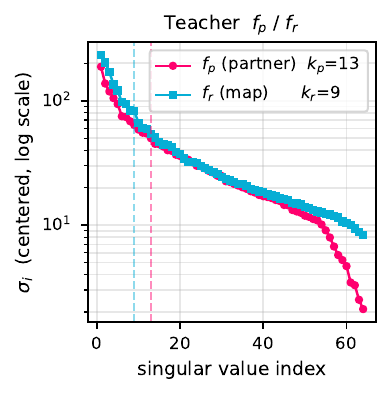}}%
\begin{minipage}[t]{0.45\linewidth}
\vspace{0pt}
\centering
\parbox[c][\ht\rankplotbox][c]{\linewidth}{\centering\usebox{\rankplotbox}}
\caption{\textbf{Singular value spectra of the teacher features.} Centered
\svd of the teacher features $\hat{f}^{p}$ (partner) and $\hat{f}^{r}$ (road) over the alignment manifest;
dashed lines mark the $80\%$ cumulative-energy index, the truncations used in
\S\ref{sec:method:alignment}.}
\label{fig:svd-latent-rank}
\end{minipage}\hfill
\begin{minipage}[t]{0.50\linewidth}
\vspace{0pt}
\centering
\captionsetup{type=table,position=below}
\parbox[c][\ht\rankplotbox][c]{\linewidth}{\centering\small
\renewcommand{\arraystretch}{1.15}%
\setlength{\tabcolsep}{6pt}%
\begin{tabular}{@{}lc@{}}
\toprule
Structural target & CL \hdscore \\
\midrule
Lower rank ($k_p{=}7,\,k_r{=}5$) & $0.517$ \\
Low-rank ($k_p{=}13,\,k_r{=}9$, headline) & \textbf{0.548} \\
Higher rank ($k_p{=}26,\,k_r{=}18$) & $0.533$ \\
Full coordinates & $0.456$ \\
\bottomrule
\end{tabular}}
\captionof{table}{\textbf{Rank ablation.} Closed-loop \hdscore{} on all $436$ \hugsim{}
scenarios. All rows use the headline data, including the $300$K \hugsim{}
domain-adaptation frames.}
\label{tab:rank-ablation}
\end{minipage}
\end{figure}

\paragraph{The teacher's features are low-rank.}
\label{sec:exp-rank}
Figure~\ref{fig:svd-latent-rank} shows where the teacher's structure lives.
The road spectrum concentrates $80\%$ of its energy in the first $9$
directions, the partner spectrum in the first $13$ of $64$. The energy of
direction $i$ is its squared singular value $\sigma_i^2$, so the cumulative
fraction $\sum_{j\le i}\sigma_j^2 / \sum_j\sigma_j^2$ is the variance
explained by the first $i$ directions. The structural loss therefore
matches pairwise similarities only within these dominant subspaces, and
the cutoffs are read from the spectrum rather than tuned. We expect two
failure modes when the truncation moves either way. A rank set too low
discards directions that carry real structure, so the loss stops
constraining the student where constraint matters. A rank set too high
pulls in the weakly structured tail, whose variation the image cannot
explain, so the encoder is dragged toward a target it cannot identify.
The ablation in Table~\ref{tab:rank-ablation} bears both
expectations out: performance peaks at the spectrum-predicted cutoff and
drops on either side, most at full coordinates.

\paragraph{The losses are nearly lossless.}
The two losses are the only channel through which the teacher reaches the
student. If they failed to express some of the teacher's competence, no
encoder, however perfect, could recover the missing part. We therefore
first quantify the cost of the objective itself. A \emph{privileged probe}
provides the measurement. The probe replaces the vision encoders with
freshly initialized encoders of the teacher's own architecture, the two
DeepSets of \S\ref{sec:method:selfplay}, which consume the privileged
vector observation $o_t$ instead of images. We train it with the same
losses on the same \nuplan{} paired frames, without the \hugsim{}
domain-adaptation frames, and the frozen head and the teacher's
structural definition remain untouched. The probe therefore differs from
the teacher in one variable only: its encoders are trained, not inherited.
Two paired comparisons read the result. On the \nuscenes{} split, the
probe reaches $0.645$ CL \hdscore{} against the teacher's $0.647$:
with identical vector inputs, the gap of $0.002$ is negligible, so the
alignment objective itself costs almost nothing. The camera
student trained on the same \nuplan{} frames
(Table~\ref{tab:paradigm-ablation}) reaches $0.574$ against the probe's
$0.645$: with an identical objective and identical data, the gap of $0.071$
combines the cost of perception with the \nuplan-to-\hugsim{} domain gap. Adding
the $300$K \hugsim{} domain-adaptation frames raises the headline student to
$0.636$, within $0.009$ of the probe. That residual perception
gap is the asymmetric-observation
problem LEAD~\citep{nguyen2025lead} studies: the teacher decides from
vectorized state that can include agents the camera cannot see.

\begin{table}[t]
\caption{\textbf{Paradigm ablation under matched data and compute.}
Closed-loop \hdscore{} on the \nuscenes{} split. Each row uses the same
$1.83$M \nuplan{} paired frames without the \hugsim{}
domain-adaptation frames and the same compute budget; the cascade replaces the
vision encoders with a SparseDrive perception module. The second block holds out
one alignment loss term at a time, trained on the same data and compute
budget.}
\label{tab:paradigm-ablation}
\centering
\small
\renewcommand{\arraystretch}{1.15}%
\setlength{\tabcolsep}{6pt}%
\begin{tabular}{@{}lc@{}}
\toprule
Training paradigm & CL \hdscore \\
\midrule
Imitation (\nuplan{} logged actions) & $0.206$ \\
Modular cascade (SparseDrive) & $0.361$ \\
Self-play pretraining + alignment (ours) & \textbf{0.574} \\
\midrule
Action-only distillation & $0.434$ \\
Structural-only distillation & $0.373$ \\
\bottomrule
\end{tabular}
\end{table}

\paragraph{The training paradigm sets the score.}
The first block of Table~\ref{tab:paradigm-ablation} measures how the choice
of training paradigm shapes the camera student, evaluated closed-loop on
the \nuscenes{} split: each row holds the
$1.83$M \nuplan{} paired frames without the \hugsim{}
domain-adaptation frames, the perception labels, and the compute budget
fixed, and varies only the supervision source. The cascade row recovers
the teacher's vector inputs from the image with a SparseDrive perception
module~\citep{sun2024sparsedrive} that we finetune under the same compute
budget, and feeds the recovered inputs to the frozen head. The cascade lands at
$0.361$. Our self-play alignment reaches $0.574$. \nuplan-action
imitation, trained on the same paired frames, scores $0.206$: the
training paradigm alone moves the student from $0.206$ to $0.574$. The
training paradigm matters.

\paragraph{Both losses are necessary.}
The second block of Table~\ref{tab:paradigm-ablation} holds out one
alignment loss at a time. Action-only distillation reaches $0.434$,
structural-only $0.373$, and both fall well short of the combined
$0.574$: each loss secures its own part of the teacher, the action
distribution or the scene relations, and neither part suffices alone.

%% file: sections/2_related.tex
% !TeX root = ../main.tex

\section{Related Work}
\label{sec:related}

Most end-to-end driving policies imitate logged human driving~\citep{codevilla2019cilrs,uniad,vad}.
However, driving competence can also be learned without human demonstrations through multi-agent self-play in vector-state simulators such as GigaFlow~\citep{cusumano2025robust}, TerraZero~\citep{wu2026terrazero}, and PufferDrive~\citep{pufferdrive2025github}.
This raises the question of how to bring the resulting competence to camera-based policies.

Several systems pursue this goal along two routes.
The first route learns to drive in vector state and then transfers the behavior to images.
Concurrent with our work, GigaPixel~\citep{rowe2026gigapixel} distills its vector-state teacher into a pixel-based student through DAgger in a bounding-box renderer, then freezes that student's head to adapt perception to real images.
DriveZero~\citep{he2026drivezero} instead trains a separate student planner, supervised by trajectories that a PPO teacher rolls out from logged frames.
Although their transfer procedures differ, neither reuses the teacher's policy head.
The second route learns the behavior directly in image space, skipping the vector-state teacher.
Pictura~\citep{yin2026pictura}, also concurrent, renders each agent's egocentric view at every self-play step and trains policies from these non-photorealistic perspective images.
Cheap rendering scales the interaction, but the resulting policies remain in the renderer's domain; a policy that consumes real camera images still requires a further training stage.
Photorealistic 3D Gaussian splatting simulators close this visual gap, but rendering every rollout at photorealistic quality is costly, so works like RAD~\citep{gao2025rad} and World Engine~\citep{li2026worldengine} use them only to post-train imitation-pretrained image policies rather than to learn driving from scratch.

We take the first route, but retain the policy head learned through vector-state self-play and train encoders that produce its scene features from images instead of vector state.
RAP~\citep{feng2025rap} likewise pairs lightweight rasterization with feature alignment, but applies them to imitation learning: it rasterizes counterfactual views from annotated logs and aligns raster features to real ones, whereas our alignment transfers the decisions of a self-play teacher.
Concretely, on fixed paired (image, scene-state) frames we compute action-distribution targets per state and feature-relation targets across the batch, and, for the planning student, plan targets from the frozen trajectory head.
Because these frames are fixed, alignment requires neither online rendering nor teacher rollouts.

%% file: sections/6_conclusion.tex
% !TeX root = ../main.tex

\section{Conclusion and Future Work}
\label{sec:conclusion}

Our closed-loop and planning results support a simple three-stage recipe:
learn driving behavior through multi-agent self-play, distill it into a
trajectory teacher that also plans, and align vision students to either
teacher for control and for planning. We train a self-play policy in
\terrazero, our vectorized simulator, freeze it as the teacher, distill its
rollouts into a trajectory teacher, and train vision frontends that produce
each teacher's scene features from paired (image, scene-state) frames. No
stage trains against logged ego trajectories: the teachers' behavior comes
from self-play, and the alignment targets come from the frozen heads the
students reuse.

In the tested scenarios, the teacher outperforms imitation-trained
pipelines, and the aligned vision policy retains most of that
competence. The same teacher also yields a planner: the trajectory student
reaches $59.4$ EPDMS on NAVSIM \texttt{navhard}, above every published
method, and $95.5$ PDMS on \texttt{navtest}, within $0.4$ points of the
privileged best-of-$16$ trajectory teacher. Imitation can only replay maneuvers that a fleet log
recorded, so its coverage of rare events is capped by what happened to
occur; self-play generates collisions, recoveries, and negotiations on
demand. We still use what the fleet log sees, but no longer what it
does. In our matched-data comparison, self-play pretraining is therefore a stronger alternative to imitation-based
pretraining for learning driving behavior, while imitation remains useful
for visual cues unavailable in bounding-box state, such as brake lights
and hand signals.

Two extensions follow directly from this design. The teacher may base
decisions on occluded agents that the camera cannot observe. Paired-frame
alignment does not remove this gap; the privileged probe bounds the total
perception cost (\S\ref{sec:exp-ablations}), but we have not separated the
occlusion component from other perception errors.
LEAD~\citep{nguyen2025lead} studies the asymmetry directly,
Pictura~\citep{yin2026pictura} sidesteps it by learning to drive from
the rendered image itself, and belief-aware privileged
distillation~\citep{haklidir2026belief} adapts the teacher's guidance to
what the student can observe. It is natural to extend the self-play training and distillation to these methods.
Alignment also draws from \nuplan{} logs and reconstructed scenes, so it
rarely encounters severe weather, unusual times of day, or rare scenario
geometries. Controllable scene generation offers a way out: video world
models such as DriveDreamer~\citep{gao2023drivedreamer} generate the
image side of our pairs, Scenario Dreamer~\citep{rowe2025scenariodreamer}
produces the scene-state side as vectorized latents, and
InfiniCube~\citep{lu2025infinicube} and SCube~\citep{huang2024scube}
synthesize street scenes beyond the recorded weather and time of day.
Pairing such generators with the self-play teacher would let both sides
of the alignment come from procedurally controlled worlds.

%% file: sections/supplementary.tex
% !TeX root = ../main.tex

\clearpage

\section{Teacher and Simulator Details}
\label{sec:supp-teacher-simulator}
\subsection{Self-Play Policy Details}
\label{sec:supp-self-play-details}

\subsubsection{Training Details}
\label{sec:supp-training-details}

\paragraph{Environment setup.}
Self-play training runs in \terrazero, the TerraZero vectorized driving
simulator, and uses \nuplan HD maps as map geometry only. We compile
\nuplan map geometry into simulator binaries and use lane topology, lane
boundaries, and drivable-area polygons as static scene context; we do
\emph{not} use \nuplan logged trajectories, logged agent initial states,
logged goals, human actions, or replayed traffic as training supervision.
Controlled vehicles are spawned procedurally from lane centerlines by
rejection sampling, so the initial scene is collision-free and drivable. The
learned policy controls vehicles only; the
simulator also procedurally places vulnerable road users and static obstacles
as scene context, but these entities are not trained policies. Goals are also
procedural. For each spawned vehicle, the simulator samples a target distance
between $20$ and $100$\,m and walks forward along the HD-map lane graph,
selecting reachable lane successors and filtering candidates by connectivity
and heading consistency. When a vehicle reaches its current goal, a new
reachable goal is sampled from the same topology; no destination is taken from
a log.

\paragraph{RL algorithm.}
The teacher is trained as a decentralized policy with PPO. We use
V-trace-corrected GAE ($\gamma=0.99$, $\lambda_{\text{GAE}}=0.95$), clipped PPO with clip
coefficient $0.2$, entropy coefficient $0.01$, value coefficient $0.5$, Adam
with an annealed $5{\times}10^{-4}$ learning rate, and \texttt{bfloat16} mixed
precision.
The reference teacher run was trained on $16$ A100 GPUs, with approximately $3$M cumulative agent steps/s throughput.

\subsubsection{Observations}
\label{sec:supp-observations}

At each simulator step ($\Delta t=0.1$\,s), a controlled vehicle observes
three groups of inputs matching the policy factorization in
Eq.~\ref{eq:encoders}: an ego vector, a variable-size set of road elements,
and a variable-size set of nearby partners.
The ego vector contains the vehicle kinematic state
$(x,y,\theta,v,a_{\text{long}},a_{\text{lat}},\phi)$ in the ego frame,
the current navigation/goal signal, the sampled vehicle geometry and
dynamics coefficients, and the per-episode reward-weight vector.
Road observations encode nearby lane and boundary polylines and traffic
controls (traffic lights and stop signs) relative to the ego frame. Partner observations encode neighboring scene entities by relative
pose, velocity, heading, size, type, and validity masks. The road and partner
sets are pooled by the DeepSets encoders described in
\S\ref{sec:method:selfplay}.

\subsubsection{Reward Components}
\label{sec:supp-rewards}

The vehicle reward in Eq.~\ref{eq:reward} uses most of the reward components
and settings from the self-play setup of \citet{cusumano2025robust}, including
the per-agent, per-episode reward-weight randomization exposed through the ego
observation. Our only reward modification is the addition of a lane-change
reward term that penalizes excessive lane changes during self-play.
Table~\ref{tab:supp-vehicle-reward-coefficients} lists the exact coefficient
settings used in our training run.

\begin{table}[htbp]
\centering
\caption{Vehicle reward coefficients used during self-play. Ranges denote
independent per-vehicle, per-episode uniform sampling.}
\label{tab:supp-vehicle-reward-coefficients}
\footnotesize
\setlength{\tabcolsep}{3pt}
\begin{tabular}{@{}p{0.22\linewidth}p{0.39\linewidth}p{0.31\linewidth}@{}}
\toprule
Reward parameter & Coefficient setting & Effect \\
\midrule
Collision & $\alpha_{\text{collision}}\sim\mathcal{U}(0.0,3.0)$ & Collision avoidance \\
Collision speed & $\alpha_{\text{coll-speed}}=0.1$ & Impact severity \\
Boundary & $\alpha_{\text{boundary}}\sim\mathcal{U}(0.0,3.0)$ & Drivable-area compliance \\
Comfort & $\alpha_{\text{comfort}}\sim\mathcal{U}(0.0,0.1)$ & Acceleration/jerk comfort \\
Lane alignment & $\alpha_{l\text{-align}}\sim\mathcal{U}(2.5{\times}10^{-4},2.5{\times}10^{-2})$ & Lane heading \\
Velocity alignment & $\alpha_{v\text{-align}}\sim\mathcal{U}(0.0,1.0)$ & Forward progress \\
Lane centering & $\alpha_{l\text{-center}}\sim\mathcal{U}(2.5{\times}10^{-4},7.5{\times}10^{-3})$ & Lane position \\
Reverse & $\alpha_{\text{reverse}}\sim\mathcal{U}(2.5{\times}10^{-4},7.5{\times}10^{-3})$ & Reverse motion \\
Lane change & $\alpha_{\text{lane-change}}\sim\mathcal{U}(0.0,0.1)$ & Lane-change regularization \\
Velocity & $\alpha_{\text{velocity}}=2.5{\times}10^{-3}$ & Speed shaping \\
Timestep & $\alpha_{\text{timestep}}=2.5{\times}10^{-5}$ & Per-step cost \\
Goal tolerance & $\delta_{\text{goal}}\sim\mathcal{U}(2.0,12.0)$\,m & Completion radius \\
Goal & $\alpha_{\text{goal}}=1.0$ & Goal completion \\
\bottomrule
\end{tabular}
\end{table}

\subsubsection{Vehicle Dynamics}
\label{sec:supp-vehicle-dynamics}

We use the jerk-actuated bicycle dynamics and the discrete $12$-action
longitudinal/lateral jerk grid from the vehicle branch of
\citet{cusumano2025robust}. As in that setup, each controlled vehicle
maintains pose, speed, longitudinal and lateral acceleration, and steering
state; selected jerk actions are integrated into accelerations, speed,
steering, and finally pose using the simulator time step.

Our dynamics setup differs from \citet{cusumano2025robust} in three ways.
First, because the learned policy controls vehicles only, dynamics
randomization is restricted to passenger-car-sized bodies rather than the
broader road-user size range used in that work:
\begin{equation}
\textit{length}\sim\mathcal{U}(3.5,5.5)\,\mathrm{m},\qquad
\textit{width}\sim\mathcal{U}(1.5,2.5)\,\mathrm{m},\qquad
\textit{height}\sim\mathcal{U}(1.2,2.0)\,\mathrm{m}.
\label{eq:supp-size-randomization}
\end{equation}
Second, at every episode reset each vehicle samples the scalar dynamics
coefficients
\begin{equation}
C_{\text{throttle}},C_{\text{steer}},C_{\text{acc}},C_{\text{vel}}
\sim\mathcal{U}(0.5,1.5),
\label{eq:supp-dyn-randomization}
\end{equation}
rather than the mixed-uniform coefficient distributions in
\citet{cusumano2025robust}. These coefficients and the sampled vehicle
dimensions are included in the ego observation so the policy can condition
on the vehicle it is controlling. Third, we run self-play with simulator
step $\Delta t=0.1$\,s, which improves closed-loop reactivity and matches the
control rate used by the student vision policy.

\subsubsection{From Jerk Actions to Simulator State}
\label{sec:supp-dynamics-update}

The policy's action at step $t$ is a pair of longitudinal and lateral jerk commands $(j^{\parallel}_t, j^{\perp}_t)$ drawn from the discrete $12$-action grid of \S\ref{sec:supp-vehicle-dynamics}. The simulator integrates this command into the bicycle-model state in one causal closed-form pass over the simulator step $\Delta t$: accelerations first, then speed, then steering, then pose, each update consuming only quantities computed earlier in the same pass.

Writing $L$ for the wheelbase of the sampled vehicle, the two accelerations accumulate the commanded jerk,
\begin{equation}
a^{\parallel}_{t+1} = a^{\parallel}_t + j^{\parallel}_t\,\Delta t,
\qquad
a^{\perp}_{t+1} = a^{\perp}_t + j^{\perp}_t\,\Delta t,
\label{eq:supp-jerk-acc}
\end{equation}
the speed integrates the updated longitudinal acceleration, and the steering angle $\phi$ carried in the ego state is the angle whose bicycle relation realizes the updated lateral acceleration at the updated speed:
\begin{equation}
v_{t+1} = v_t + a^{\parallel}_{t+1}\,\Delta t,
\qquad
\phi_{t+1} = \arctan\!\Big(\frac{a^{\perp}_{t+1}\,L}{v_{t+1}^{2}}\Big).
\label{eq:supp-jerk-speed-steer}
\end{equation}
The pose finally integrates the updated speed and the yaw rate $\omega_{t+1} = v_{t+1}\tan\phi_{t+1}/L = a^{\perp}_{t+1}/v_{t+1}$ implied by the bicycle model:
\begin{equation}
x_{t+1} = x_t + v_{t+1}\cos\theta_t\,\Delta t, \quad
y_{t+1} = y_t + v_{t+1}\sin\theta_t\,\Delta t, \quad
\theta_{t+1} = \theta_t + \omega_{t+1}\,\Delta t.
\label{eq:supp-jerk-pose}
\end{equation}

These are the updates the teacher trains under. At \hugsim evaluation the same integrator consumes each action and hands the resulting state to the simulator, so no second dynamics model and no sim-to-sim conversion sit between the policy and the benchmark.

\subsection{Closed-Loop HD-Score}
\label{sec:supp-cl-hdscore}

The native HUGSim~\citep{hugsim} HD-Score is not well designed for closed-loop evaluation: it scores the policy's \emph{planned} future trajectory rather than the path the ego actually drives.\footnote{Throughout this appendix, the \emph{native} HD-Score denotes the public HUGSim scorer \texttt{sim/utils/score\_calculator.py} (repository \texttt{hyzhou404/HUGSIM}, commit \texttt{62c690d}, 2025-11-08; \url{https://github.com/hyzhou404/HUGSIM}), whose \texttt{calculate()} routine is the scoring entry point we benchmark against; all statements about native behavior below are taken from it.}
This is a legacy of the open-loop NAVSIM-style protocol~\citep{navsim} it was written for, where each keyframe carries a multi-second planned trajectory and the metric never has to commit to what is executed.
But what we want to measure is closed-loop driving quality, which lives in the \emph{realized} rollout trajectory, not in a plan the policy may never follow---and our single-step policy emits no multi-second plan to begin with.
We therefore introduce a closed-loop variant of the HD-Score (used in \S\ref{sec:exp-setup}) that scores the realized rollout directly: we keep the aggregation formula, the no-collision (NC) and drivable-area-compliance (DAC) subscores, and the route-completion term ($R_c$) unchanged, and redefine the time-to-collision (TTC) and comfort (COM) terms to operate on the realized trajectory.

\subsubsection{Closed-Loop Redefinitions}
\label{sec:supp-cl-hdscore-defs}

Let $\mathcal{T}=\{t_1,\dots,t_M\}$ be the set of evaluation keyframes (a fixed-stride decimation of the rollout, default $0.5\,\mathrm{s}$), let $(x^e_t,y^e_t,\theta^e_t,v^e_t)$ be the realized pose and instantaneous velocity of the ego at keyframe $t$, and let $\mathcal{P}_t$ index the set of valid vehicle-like partners visible at keyframe $t$.
We continue to compute the aggregate as
\begin{equation}
\mathrm{HD\text{-}Score}\;=\;\frac{1}{|\mathcal{T}|}\sum_{t\in\mathcal{T}}\mathrm{NC}_t\cdot\mathrm{DAC}_t\cdot\frac{5\,\mathrm{TTC}_t+2\,\mathrm{COM}_t}{7}\;\cdot\;R_c,
\label{eq:supp-cl-hdscore-agg}
\end{equation}
with NC, DAC, and $R_c$ defined identically to native HUGSim and TTC, COM redefined below.

\paragraph{Closed-loop TTC.}
Native TTC has three issues for closed-loop use: (i) it probes the \emph{planned} trajectory, not the realized one; (ii) it is binary and checks only two shifts, $\Delta\in\{0.5,1.0\}\,\mathrm{s}$, so it cannot grade how imminent a conflict is; and (iii) it tests against \emph{frozen} partners, ignoring their motion.
We discard the planned trajectory and measure imminent risk directly on the realized rollout, addressing all three issues above.
At keyframe $t$ we propagate the ego \emph{and} every valid partner $j\in\mathcal{P}_t$ forward by their observed instantaneous velocities along a short horizon $H_{\text{TTC}}$ (default $1.0\,\mathrm{s}$, the conventional imminent window and the native velocity-shift upper bound) at a fine stride $\Delta_{\text{TTC}}$ (default $0.2\,\mathrm{s}$):
\begin{equation}
(x^{(a)}_{t,\delta},\,y^{(a)}_{t,\delta}) = (x^{(a)}_t,\,y^{(a)}_t) + \delta\,(v^{(a)}_{x,t},\,v^{(a)}_{y,t}),\quad a\in\{\mathrm{ego}\}\cup\mathcal{P}_t,\ \ \delta\in\{0,\Delta_{\text{TTC}},\ldots,H_{\text{TTC}}\}.
\label{eq:supp-cl-ttc-prop}
\end{equation}
At each $\delta$ we test the propagated ego oriented bounding box (OBB) against both each propagated partner box and the static 3D Gaussian-splatting (3DGS) point cloud $\mathcal{B}$ reconstructed for the scene background, the latter flagged when the ego box encloses more than $N_{\text{bg}}$ such points (default $100$, matching the native background-collision test), and record the earliest collision time
\begin{equation}
\tau^\star_t \;=\; \min\!\left\{\delta\;:\;\Big(\exists\,j\in\mathcal{P}_t,\;\mathrm{OBB}_t^{\text{ego}}(\delta)\cap\mathrm{OBB}_t^{(j)}(\delta)\neq\varnothing\Big)\;\vee\;\Big(\big|\mathcal{B}\cap\mathrm{OBB}_t^{\text{ego}}(\delta)\big|> N_{\text{bg}}\Big)\right\},
\label{eq:supp-cl-ttc-tau}
\end{equation}
with $\tau^\star_t = +\infty$ if no collision occurs over the horizon, and report the fractional score
\begin{equation}
\mathrm{TTC}_t \;=\; \min\!\Big(1,\,\tau^\star_t\,/\,H_{\text{TTC}}\Big)\in[0,1].
\label{eq:supp-cl-ttc}
\end{equation}
These three changes---realized poses, a fractional imminent-window sweep, and propagated (rather than frozen) partners---resolve issues~(i)--(iii) respectively.
A $\delta=0$ collision (ego already in contact at the keyframe) returns $\mathrm{TTC}_t = 0$ directly.

\paragraph{Closed-loop COM.}
We retain the native comfort bounds (longitudinal acceleration, longitudinal jerk, yaw acceleration, and yaw rate; Eq.~\ref{eq:supp-cl-com}) but apply them to the \emph{realized} ego trajectory over a fixed-duration forward window of $W_{\text{COM}}$ seconds (default $1.0\,\mathrm{s}$) starting at each keyframe.
Concretely, at keyframe $t$ we take the forward window $\mathcal{W}_t=\{t,\,t+\Delta_{\text{sim}},\,\ldots,\,t+(n-1)\Delta_{\text{sim}}\}$ of the next $n=\max(\mathrm{round}(W_{\text{COM}}/\Delta_{\text{sim}}),\,4)$ realized ego poses ($n=10$ at the defaults $W_{\text{COM}}=1.0\,\mathrm{s}$, $\Delta_{\text{sim}}=0.1\,\mathrm{s}$, with $\Delta_{\text{sim}}$ the simulator step), compute first- and second-order finite differences to obtain longitudinal and yaw kinematics, and set $\mathrm{COM}_t = 1$ iff all of these bounds hold at every sample in $\mathcal{W}_t$ (defaulting to $1$ if fewer than four samples remain, as in native), else $\mathrm{COM}_t = 0$:
\begin{equation}
\mathrm{COM}_t \;=\;\bigwedge_{s\in\mathcal{W}_t}\Big[\,a^\parallel_s\in[-4.05, 2.40]\,\wedge\,|j^\parallel_s|\le 8.37\,\wedge\,|\alpha_s|\le 1.93\,\wedge\,|\omega_s|\le 0.95\,\Big],
\label{eq:supp-cl-com}
\end{equation}
where $a^\parallel_s$ is longitudinal acceleration ($\mathrm{m/s}^2$), $j^\parallel_s$ longitudinal jerk ($\mathrm{m/s}^3$), $\alpha_s$ yaw acceleration ($\mathrm{rad/s}^2$), and $\omega_s$ yaw rate ($\mathrm{rad/s}$).
Parameterizing the window in seconds rather than step count keeps the metric invariant to the rollout $\Delta_{\text{sim}}$.

\paragraph{Route completion.}
We retain the native $R_c$ unchanged so that every row of Table~\ref{tab:hugsim-all-datasets}---published baselines, \eco, and the teacher---shares one route-completion protocol and remains comparable with the numbers HUGSim publishes.
Each HUGSim scene supplies the camera-pose sequence $(\mathbf{r}_0,\dots,\mathbf{r}_{N-1})\subset\mathbb{R}^2$ of the reconstructed capture drive as its reference route.
At each simulator step $t$ the ego position is snapped to its nearest camera pose $k_t=\arg\min_{k}\,\|(x^e_t,y^e_t)-\mathbf{r}_k\|$ and the per-step completion is the index fraction $r_t=(k_t+1)/(0.9N)$; the scenario-level score is
\begin{equation}
R_c \;=\; \min\!\Big(1,\;\max_{t}\,r_t\Big).
\label{eq:supp-cl-rc}
\end{equation}
Two properties of this definition matter when reading the tables. It is an index fraction rather than a physical distance, so it depends on the pose spacing of the capture drive, and the $0.9$ normalizer saturates $R_c$ at one once the ego reaches $90\%$ of the poses. $R_c=1$ therefore means the ego reached the final tenth of the reference route, not necessarily its endpoint. Both properties apply equally to every method in our tables and to HUGSim's published baselines.

\subsubsection{Native vs.\ Closed-Loop Summary}
\label{sec:supp-cl-hdscore-table}

\begin{table}[htbp]
    \centering
    \caption{Native HD-Score vs.\ closed-loop HD-Score per subscore. NC, DAC, $R_c$, and the aggregation formula are unchanged; TTC and COM are redefined to operate on the realized rollout.}
    \label{tab:supp-cl-hdscore-comparison}
    \scriptsize
    \renewcommand{\arraystretch}{1.25}
    \setlength{\tabcolsep}{4pt}
    \begin{tabular}{@{}p{0.10\linewidth}p{0.40\linewidth}p{0.40\linewidth}@{}}
    \toprule
    Sub & Native HUGSim HD-Score & Closed-loop variant (this work) \\
    \midrule
    NC  & OBB ego--partner collision per keyframe; vehicle-only partner filter; output $\in\{0,1\}$. & Same. \\
    DAC & Four-corner-inside check against per-scene drivable-area polylines; output $\in\{0, 0.5, 1\}$. & Same. \\
    TTC & Entire planned trajectory ($7$ poses, $\le\!3.5\,\mathrm{s}$) rigidly shifted by ego velocity at two horizons $\Delta\in\{0.5,1.0\}\,\mathrm{s}$; collision of every shifted ego pose vs.\ \emph{frozen} partners $+$ scene geometry ($14$ probes); binary $\{0,1\}$. & Realized ego \emph{and} all valid partners velocity-propagated on a fine grid (stride $\Delta_{\text{TTC}}=0.2\,\mathrm{s}$, horizon $H_{\text{TTC}}=1.0\,\mathrm{s}$); earliest ego collision vs.\ propagated partners $+$ 3DGS background ($>\!N_{\text{bg}}$ enclosed points) gives fractional $\min(1,\,\tau^\star_t/H_{\text{TTC}})\in[0,1]$ (Eq.~\ref{eq:supp-cl-ttc}); $\delta{=}0$ collision $\Rightarrow$ TTC$_t = 0$. \\
    COM & Kinematic comfort bounds (lon.\ accel., lon.\ jerk, yaw accel., yaw rate; a lateral bound is coded but inert) on the keyframe's \emph{planned} trajectory poses; binary $\{0,1\}$. & Same active bounds on the \emph{realized} ego trajectory over a seconds-parameterized window of duration $W_{\text{COM}}=1.0\,\mathrm{s}$ (Eq.~\ref{eq:supp-cl-com}). \\
    $R_c$ & $\min(1, \max_t r_t)$ with $r_t=(k_t{+}1)/(0.9N)$: a camera-pose-\emph{index} fraction ($k_t$ = nearest pose index, $N$ = pose count) that saturates at $90\%$ of the poses. & Same (Eq.~\ref{eq:supp-cl-rc}); retained so all rows share one route-completion protocol. \\
    Agg.\ & $\mathrm{mean}_t[\mathrm{NC}_t\cdot\mathrm{DAC}_t\cdot(5\,\mathrm{TTC}_t+2\,\mathrm{COM}_t)/7]\cdot R_c$. & Same (Eq.~\ref{eq:supp-cl-hdscore-agg}). \\
    \bottomrule
    \end{tabular}
\end{table}

Because the redefined TTC and COM read different motion from native (realized rollout vs.\ planned trajectory) and, for TTC, a different scale (continuous vs.\ binary), neither is directly comparable to its native counterpart per keyframe, even on the same rollout. NC and DAC are comparable. The detailed comparison is summarized in Table~\ref{tab:supp-cl-hdscore-comparison}. 

\medskip
Table~\ref{tab:supp-cl-hdscore-comparison} defines the two scorers. Every row of Table~\ref{tab:hugsim-all-datasets}, baselines included, is scored with the closed-loop variant in its left block, applied identically to all methods: each policy re-plans at every simulator step, executes only its first action, and the scorer reads the single realized rollout. On \nuscenes{}, native scoring of the same rollouts preserves the closed-loop ranking ECO $>$ LTF $>$ \{VAD, UniAD\} (VAD and UniAD differ by at most $0.008$ under either scorer), so the headline ordering does not depend on which variant is used; per-subscore values for the teacher and the student appear in Tables~\ref{tab:teacher-hugsim-full} and~\ref{tab:supp-hugsim-ours}.

\subsection{Self-Play Policy Baseline Performance}
\label{sec:supp-self-play-policy-baseline}

See \S\ref{sec:exp-teacher} for the teacher's planning-benchmark strength and cross-map transfer. Table~\ref{tab:teacher-hugsim-full} reports the teacher, and Table~\ref{tab:supp-hugsim-ours} reports the aligned student, across all $436$ \hugsim{} scenarios under both scorers.

\begin{table*}[htbp]
    \centering
    \caption{Self-play teacher across all $436$ HUGSim scenarios and difficulty
    tiers, under both scorers. The teacher uses privileged vectorized scene
    state; background traffic is replayed. The Native block scores the decel
    stand-in plan with HUGSim's planned-trajectory subscores; the Closed-loop
    block scores the realized rollout with the redefined TTC and COM. $R_c$ is
    shared. PDMS is the NC, DAC, TTC, and COM aggregate before route
    completion (HD-PDMS); HD $=$ PDMS $\times R_c$.}
    \label{tab:teacher-hugsim-full}
    \scriptsize
    \renewcommand{\arraystretch}{1.0}
    \setlength{\tabcolsep}{2pt}
    \begin{tabular}{@{}lrr @{\hspace{4pt}}|@{\hspace{4pt}} rrrrrr @{\hspace{4pt}}|@{\hspace{4pt}} rrrrrr@{}}
    \toprule
     & & & \multicolumn{6}{c}{Native (decel stand-in plan)} & \multicolumn{6}{c}{Closed-loop (realized rollout)} \\
    \cmidrule(lr){4-9}\cmidrule(lr){10-15}
    Set & N & $R_c$ & NC & DAC & TTC & COM & PDMS & HD & NC & DAC & TTC & COM & PDMS & HD \\
    \midrule
    \multicolumn{15}{@{}l}{\textit{\nuscenes}} \\
    \quad Easy & 18 & 0.949 & 0.944 & 0.997 & 0.934 & 1.000 & 0.925 & 0.875 & 0.980 & 0.999 & 0.967 & 0.978 & 0.954 & 0.906 \\
    \quad Medium & 34 & 0.716 & 0.923 & 0.999 & 0.907 & 1.000 & 0.903 & 0.657 & 0.978 & 1.000 & 0.964 & 0.971 & 0.950 & 0.684 \\
    \quad Hard & 18 & 0.732 & 0.832 & 0.997 & 0.835 & 1.000 & 0.817 & 0.653 & 0.965 & 0.999 & 0.942 & 0.947 & 0.920 & 0.685 \\
    \quad Extreme & 18 & 0.301 & 0.743 & 0.997 & 0.713 & 1.000 & 0.707 & 0.250 & 0.959 & 0.999 & 0.925 & 0.931 & 0.889 & 0.279 \\
    \quad \textbf{All} & 88 & 0.682 & 0.872 & 0.998 & 0.858 & 1.000 & 0.850 & 0.618 & 0.972 & 0.999 & 0.952 & 0.959 & 0.932 & 0.647 \\
    \midrule
    \multicolumn{15}{@{}l}{\textit{Waymo}} \\
    \quad Easy & 22 & 0.959 & 0.966 & 0.995 & 0.963 & 1.000 & 0.953 & 0.912 & 0.983 & 0.995 & 0.975 & 0.993 & 0.965 & 0.924 \\
    \quad Medium & 42 & 0.870 & 0.986 & 0.976 & 0.984 & 1.000 & 0.959 & 0.847 & 0.994 & 0.977 & 0.991 & 0.999 & 0.966 & 0.854 \\
    \quad Hard & 22 & 0.860 & 0.895 & 0.995 & 0.896 & 1.000 & 0.884 & 0.777 & 0.974 & 0.995 & 0.964 & 0.996 & 0.953 & 0.817 \\
    \quad Extreme & 22 & 0.130 & 0.861 & 0.955 & 0.761 & 1.000 & 0.751 & 0.102 & 0.959 & 0.955 & 0.917 & 0.957 & 0.847 & 0.115 \\
    \quad \textbf{All} & 108 & 0.735 & 0.938 & 0.979 & 0.916 & 1.000 & 0.900 & 0.694 & 0.981 & 0.979 & 0.967 & 0.989 & 0.939 & 0.710 \\
    \midrule
    \multicolumn{15}{@{}l}{\textit{KITTI-360}} \\
    \quad Easy & 24 & 0.732 & 0.932 & 0.941 & 0.931 & 1.000 & 0.872 & 0.650 & 0.966 & 0.945 & 0.948 & 0.970 & 0.887 & 0.659 \\
    \quad Medium & 37 & 0.544 & 0.911 & 0.893 & 0.899 & 1.000 & 0.796 & 0.488 & 0.934 & 0.898 & 0.923 & 0.982 & 0.815 & 0.497 \\
    \quad Hard & 26 & 0.278 & 0.633 & 0.802 & 0.618 & 1.000 & 0.456 & 0.205 & 0.858 & 0.811 & 0.843 & 0.975 & 0.648 & 0.244 \\
    \quad Extreme & 26 & 0.038 & 0.655 & 0.555 & 0.594 & 1.000 & 0.264 & 0.015 & 0.823 & 0.569 & 0.797 & 0.924 & 0.354 & 0.019 \\
    \quad \textbf{All} & 113 & 0.406 & 0.793 & 0.805 & 0.771 & 1.000 & 0.611 & 0.348 & 0.898 & 0.812 & 0.881 & 0.964 & 0.686 & 0.363 \\
    \midrule
    \multicolumn{15}{@{}l}{\textit{\pandaset}} \\
    \quad Easy & 15 & 0.912 & 0.973 & 1.000 & 0.968 & 1.000 & 0.964 & 0.876 & 0.990 & 1.000 & 0.984 & 0.990 & 0.978 & 0.890 \\
    \quad Medium & 45 & 0.892 & 0.984 & 1.000 & 0.980 & 1.000 & 0.980 & 0.872 & 0.994 & 1.000 & 0.992 & 0.999 & 0.990 & 0.883 \\
    \quad Hard & 30 & 0.307 & 0.864 & 1.000 & 0.842 & 1.000 & 0.847 & 0.275 & 0.979 & 1.000 & 0.960 & 0.984 & 0.947 & 0.302 \\
    \quad Extreme & 37 & 0.232 & 0.820 & 1.000 & 0.785 & 1.000 & 0.793 & 0.186 & 0.980 & 1.000 & 0.961 & 0.985 & 0.949 & 0.224 \\
    \quad \textbf{All} & 127 & 0.564 & 0.907 & 1.000 & 0.889 & 1.000 & 0.892 & 0.532 & 0.986 & 1.000 & 0.974 & 0.990 & 0.967 & 0.554 \\
    \midrule
    \multicolumn{15}{@{}l}{\textit{All four subsets}} \\
    \quad Easy & 79 & 0.879 & 0.952 & 0.980 & 0.948 & 1.000 & 0.924 & 0.817 & 0.979 & 0.982 & 0.967 & 0.982 & 0.941 & 0.833 \\
    \quad Medium & 158 & 0.767 & 0.954 & 0.968 & 0.946 & 1.000 & 0.915 & 0.729 & 0.977 & 0.970 & 0.969 & 0.989 & 0.934 & 0.742 \\
    \quad Hard & 96 & 0.506 & 0.803 & 0.945 & 0.793 & 1.000 & 0.744 & 0.442 & 0.942 & 0.947 & 0.926 & 0.977 & 0.862 & 0.476 \\
    \quad Extreme & 103 & 0.173 & 0.774 & 0.878 & 0.719 & 1.000 & 0.636 & 0.136 & 0.932 & 0.881 & 0.904 & 0.954 & 0.767 & 0.158 \\
    \quad \textbf{All} & 436 & 0.589 & 0.878 & 0.944 & 0.859 & 1.000 & 0.813 & 0.542 & 0.959 & 0.946 & 0.944 & 0.977 & 0.880 & 0.562 \\
    \midrule
    \bottomrule
    \end{tabular}
\end{table*}

\begin{table*}[htbp]
    \centering
    \caption{Aligned end-to-end student across all $436$ HUGSim scenarios and
    difficulty tiers, under both scorers. The student consumes camera images
    and emits one jerk action per step. The Native block scores the decel
    stand-in plan (\S\ref{sec:supp-cl-hdscore}); the Closed-loop block scores
    the realized rollout. $R_c$ is shared. PDMS is the NC, DAC, TTC, and COM
    aggregate before route completion (HD-PDMS); HD $=$ PDMS $\times R_c$.}
    \label{tab:supp-hugsim-ours}
    \scriptsize
    \renewcommand{\arraystretch}{1.0}
    \setlength{\tabcolsep}{2pt}
    \begin{tabular}{@{}lrr @{\hspace{4pt}}|@{\hspace{4pt}} rrrrrr @{\hspace{4pt}}|@{\hspace{4pt}} rrrrrr@{}}
    \toprule
     & & & \multicolumn{6}{c}{Native (decel stand-in plan)} & \multicolumn{6}{c}{Closed-loop (realized rollout)} \\
    \cmidrule(lr){4-9}\cmidrule(lr){10-15}
    Set & N & $R_c$ & NC & DAC & TTC & COM & PDMS & HD & NC & DAC & TTC & COM & PDMS & HD \\
    \midrule
    \multicolumn{15}{@{}l}{\textit{\nuscenes}} \\
    \quad Easy & 18 & 0.934 & 0.957 & 1.000 & 0.950 & 1.000 & 0.946 & 0.884 & 0.981 & 1.000 & 0.972 & 0.975 & 0.962 & 0.901 \\
    \quad Medium & 34 & 0.721 & 0.916 & 0.998 & 0.900 & 1.000 & 0.897 & 0.650 & 0.974 & 1.000 & 0.956 & 0.974 & 0.943 & 0.682 \\
    \quad Hard & 18 & 0.690 & 0.853 & 0.997 & 0.850 & 1.000 & 0.838 & 0.620 & 0.963 & 0.999 & 0.946 & 0.985 & 0.934 & 0.641 \\
    \quad Extreme & 18 & 0.295 & 0.760 & 1.000 & 0.736 & 1.000 & 0.728 & 0.251 & 0.968 & 1.000 & 0.940 & 0.945 & 0.913 & 0.278 \\
    \quad \textbf{All} & 88 & 0.671 & 0.880 & 0.999 & 0.867 & 1.000 & 0.860 & 0.610 & 0.972 & 1.000 & 0.954 & 0.971 & 0.939 & 0.636 \\
    \midrule
    \multicolumn{15}{@{}l}{\textit{Waymo}} \\
    \quad Easy & 22 & 0.963 & 0.976 & 0.988 & 0.975 & 1.000 & 0.967 & 0.936 & 0.990 & 0.991 & 0.986 & 0.997 & 0.975 & 0.941 \\
    \quad Medium & 42 & 0.856 & 0.975 & 0.988 & 0.968 & 1.000 & 0.962 & 0.827 & 0.991 & 0.991 & 0.985 & 0.992 & 0.973 & 0.836 \\
    \quad Hard & 22 & 0.800 & 0.889 & 0.974 & 0.876 & 1.000 & 0.868 & 0.717 & 0.975 & 0.984 & 0.965 & 0.972 & 0.942 & 0.758 \\
    \quad Extreme & 22 & 0.131 & 0.864 & 0.955 & 0.759 & 1.000 & 0.747 & 0.100 & 0.958 & 0.955 & 0.915 & 0.966 & 0.848 & 0.116 \\
    \quad \textbf{All} & 108 & 0.719 & 0.935 & 0.978 & 0.908 & 1.000 & 0.900 & 0.679 & 0.981 & 0.982 & 0.967 & 0.983 & 0.942 & 0.695 \\
    \midrule
    \multicolumn{15}{@{}l}{\textit{\kitti}} \\
    \quad Easy & 24 & 0.614 & 0.883 & 0.935 & 0.874 & 1.000 & 0.810 & 0.535 & 0.929 & 0.936 & 0.911 & 0.972 & 0.841 & 0.550 \\
    \quad Medium & 37 & 0.501 & 0.908 & 0.829 & 0.896 & 1.000 & 0.731 & 0.450 & 0.926 & 0.836 & 0.921 & 0.972 & 0.750 & 0.457 \\
    \quad Hard & 26 & 0.241 & 0.774 & 0.715 & 0.771 & 1.000 & 0.537 & 0.195 & 0.926 & 0.734 & 0.920 & 0.972 & 0.649 & 0.219 \\
    \quad Extreme & 26 & 0.032 & 0.715 & 0.553 & 0.652 & 1.000 & 0.298 & 0.014 & 0.844 & 0.565 & 0.825 & 0.930 & 0.376 & 0.017 \\
    \quad \textbf{All} & 113 & 0.357 & 0.827 & 0.762 & 0.806 & 1.000 & 0.603 & 0.309 & 0.908 & 0.771 & 0.896 & 0.962 & 0.660 & 0.321 \\
    \midrule
    \multicolumn{15}{@{}l}{\textit{\pandaset}} \\
    \quad Easy & 15 & 0.911 & 0.976 & 1.000 & 0.972 & 1.000 & 0.970 & 0.882 & 0.991 & 1.000 & 0.984 & 1.000 & 0.981 & 0.892 \\
    \quad Medium & 45 & 0.892 & 0.971 & 1.000 & 0.964 & 1.000 & 0.962 & 0.855 & 0.989 & 1.000 & 0.981 & 0.999 & 0.979 & 0.871 \\
    \quad Hard & 30 & 0.367 & 0.859 & 1.000 & 0.815 & 1.000 & 0.825 & 0.324 & 0.974 & 1.000 & 0.951 & 0.995 & 0.941 & 0.353 \\
    \quad Extreme & 37 & 0.251 & 0.795 & 0.994 & 0.744 & 1.000 & 0.754 & 0.189 & 0.968 & 0.997 & 0.941 & 0.986 & 0.927 & 0.234 \\
    \quad \textbf{All} & 127 & 0.584 & 0.894 & 0.998 & 0.866 & 1.000 & 0.870 & 0.539 & 0.979 & 0.999 & 0.963 & 0.994 & 0.955 & 0.565 \\
    \midrule
    \multicolumn{15}{@{}l}{\textit{All four subsets}} \\
    \quad Easy & 79 & 0.841 & 0.944 & 0.977 & 0.938 & 1.000 & 0.915 & 0.792 & 0.970 & 0.978 & 0.960 & 0.985 & 0.932 & 0.804 \\
    \quad Medium & 158 & 0.754 & 0.946 & 0.956 & 0.935 & 1.000 & 0.894 & 0.709 & 0.972 & 0.959 & 0.963 & 0.985 & 0.916 & 0.724 \\
    \quad Hard & 96 & 0.493 & 0.842 & 0.916 & 0.824 & 1.000 & 0.759 & 0.435 & 0.959 & 0.924 & 0.945 & 0.982 & 0.861 & 0.463 \\
    \quad Extreme & 103 & 0.178 & 0.784 & 0.875 & 0.723 & 1.000 & 0.633 & 0.137 & 0.934 & 0.879 & 0.906 & 0.960 & 0.769 & 0.162 \\
    \quad \textbf{All} & 436 & 0.576 & 0.884 & 0.932 & 0.861 & 1.000 & 0.807 & 0.528 & 0.960 & 0.936 & 0.945 & 0.979 & 0.872 & 0.548 \\
    \bottomrule
    \end{tabular}
\end{table*}

\medskip
The pooled closed-loop \emph{All four subsets} rows of
Tables~\ref{tab:teacher-hugsim-full} and Table~\ref{tab:supp-hugsim-ours}
($0.562$ vs.\ $0.548$) correspond to the teacher and student aggregates
discussed in \S\ref{sec:exp-headline}.

\paragraph{What limits the teacher.}
On \nuscenes{}, safety, on-road compliance, and comfort hold up across all four tiers---NC stays above $0.95$, DAC is saturated near $1.0$, TTC above $0.92$, and COM above $0.90$---so HD-PDMS never drops below $0.88$. There, the HD-Score is instead driven down almost entirely by route completion: $R_c$ falls from $0.95$ on Easy to $0.30$ on Extreme, and because it multiplies the aggregate (Eq.~\ref{eq:supp-cl-hdscore-agg}), it alone pulls the Extreme HD-Score to $0.28$ despite an HD-PDMS of $0.89$. On \kitti{}, drivable-area compliance also collapses (Extreme DAC $0.569$, HD-PDMS $0.354$) and pulls the pooled Extreme HD-PDMS down to $0.767$. Even the privileged teacher cannot make progress on many Extreme scenarios, consistent with visually implausible initial configurations (e.g., other cars passing through each other) of the kind discussed in \S\ref{sec:exp-headline}.

\section{Student and Alignment Details}
\label{sec:supp-student-alignment}
\subsection{Structural Loss Rank Ablation}
\label{sec:supp-loss-ablation}

We ablate the truncation rank of the structural loss. The headline uses $k_p=13$, $k_r=9$, the $80\%$ cumulative-energy cutoffs of the teacher spectra (\S\ref{sec:exp-rank}). We compare it with half these ranks ($k_p=7$, $k_r=5$), twice these ranks ($k_p=26$, $k_r=18$), and a full-coordinate variant that skips the projection and matches the pairwise cosine-similarity matrix over all feature coordinates. Every variant keeps the rest of the recipe fixed: the same frozen teacher, the headline data including the $300$K \hugsim{} domain-adaptation frames, the action loss, $\lambda=0.5$, and the same training budget. All variants are evaluated with closed-loop \hdscore{} on all $436$ \hugsim{} scenarios. Table~\ref{tab:rank-ablation} reports the results, and \S\ref{sec:exp-rank} discusses why performance peaks at the spectrum cutoff.

\Needspace{14\baselineskip}
\subsection{Vision Backbone Ablation}
\label{sec:supp-backbone-ablation}

\begin{wrapfigure}[11]{r}{0.34\linewidth}
\vspace{-14pt}
\centering
\includegraphics[width=\linewidth]{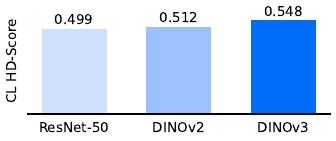}
\caption{\textbf{Vision-backbone ablation.} Closed-loop \hdscore{} under a fixed alignment recipe for each image backbone, trained on the headline data including the $300$K \hugsim{} domain-adaptation frames.}
\label{fig:backbone-ablation}
\end{wrapfigure}
We compare the selected DINOv3 backbone against the alternative image encoders DINOv2 and ResNet-50 under an identical alignment recipe: the same frozen teacher, the same paired frames, and the same adapter architecture and training budget; only the pretrained image encoder changes. The pretrained representation contributes measurably on top of the alignment objective: swapping it moves the closed-loop score from $0.499$ (ResNet-50) and $0.512$ (DINOv2) to $0.548$. The DINOv3 number matches the headline student of Table~\ref{tab:supp-hugsim-ours}; the two baselines transfer the same adapter design onto a supervised ImageNet representation (ResNet-50) and an earlier self-supervised one (DINOv2).

The ordering is consistent with what the frozen head consumes. The road and partner adapters are linear, so the head can only use the structure its input features already carry: a stronger pretrained representation enters the alignment already closer to the teacher's feature geometry, and the linear adapters have less work left to do. The absolute gaps are modest, $0.049$ between ResNet-50 and DINOv3, but they come entirely from the encoder swap, with the teacher, the data, and the recipe held fixed.

\section{NAVSIM Per-Subscore Breakdown}
\label{sec:supp-navsim-subitems}

Tables~\ref{tab:supp-navtest-subscores} and~\ref{tab:supp-navhard-subscores}
expand the NAVSIM aggregates of Table~\ref{tab:navsim} into their subscores.
All baseline entries are the published results as compiled by \citet{he2026drivezero}. On
\texttt{navtest} (NAVSIMv1), PDMS aggregates no-collision (NC), drivable-area
compliance (DAC), time-to-collision (TTC), comfort (C), and ego-progress (EP).
On \texttt{navhard} (NAVSIMv2), EPDMS aggregates NC, DAC, drivable-direction
compliance (DDC), traffic-light compliance (TLC), ego-progress (EP), TTC,
lane-keeping (LK), heading consistency (HC), and ego-centering (EC), scored
over two stages: Stage~1 scores the unperturbed scene, and Stage~2 re-scores
under Gaussian-Splatting-based ego-state perturbations. \emph{S.} is the
per-stage EPDM score. Bold marks the best result in each input group.

\begin{table}[!htb]
\centering
\captionsetup{font=normalsize}
\small
\renewcommand{\arraystretch}{1.15}
\setlength{\tabcolsep}{9pt}
    \caption{Per-subscore breakdown on the NAVSIMv1 \texttt{navtest} split.
    PDMS aggregates NC, DAC, TTC, C, and EP (one decimal). Baseline rows are
    published numbers; \textit{italicized} rows are privileged references. Bold
    marks the best result in each input group. GigaPixel~\citep{rowe2026gigapixel}
    is omitted because it reports \texttt{navhard} results only.}
    \label{tab:supp-navtest-subscores}
    \begin{tabular}{@{}lcccccc@{}}
    \toprule
    Method & NC$\uparrow$ & DAC$\uparrow$ & TTC$\uparrow$ & C$\uparrow$ & EP$\uparrow$ & PDMS$\uparrow$ \\
    \midrule
    \multicolumn{7}{@{}l}{\textit{Privileged reference (ground-truth scene state)}} \\*
    \quad \textit{Human driver} & \textbf{100} & \textbf{100} & \textbf{100} & \textbf{99.9} & 87.5 & 94.8 \\
    \quad \textit{PDM-Closed~\citep{dauner2023pdm}} & 94.6 & 99.8 & 89.9 & 86.9 & \textbf{99.9} & 89.1 \\
    \quad \textit{DriveRL~\citep{he2026drivezero}} & 99.8 & 99.9 & 99.1 & 99.0 & 91.5 & 95.8 \\
    \quad \textit{Trajectory teacher (ours)} & 99.0 & 99.6 & 93.2 & 99.8 & 97.2 & \textbf{95.9} \\
    \midrule
    \multicolumn{7}{@{}l}{\textit{Camera \& LiDAR input, human demonstrations}} \\*
    \quad TransFuser~\citep{ltf} & 97.7 & 92.8 & 92.8 & \textbf{100} & 79.2 & 84.0 \\
    \quad DRAMA~\citep{drama2025} & 98.0 & 93.1 & 94.8 & \textbf{100} & 80.1 & 85.5 \\
    \quad VADv2~\citep{vadv2} & 98.1 & 94.8 & 94.3 & \textbf{100} & 80.6 & 86.2 \\
    \quad DiffusionDrive~\citep{diffusiondrive} & 98.2 & 96.2 & 94.7 & \textbf{100} & 82.2 & 88.1 \\
    \quad Hydra-MDP++~\citep{hydramdp} & 98.6 & 98.6 & 95.1 & \textbf{100} & 85.7 & 91.0 \\
    \quad Centaur~\citep{centaur2025} & \textbf{99.5} & \textbf{98.9} & \textbf{98.0} & \textbf{100} & 85.9 & 92.6 \\*
    \quad DriveSuprim~\citep{drivesuprim} & 98.6 & 98.6 & 95.5 & \textbf{100} & \textbf{91.3} & \textbf{93.5} \\
    \midrule
    \multicolumn{7}{@{}l}{\textit{Camera input, human demonstrations}} \\*
    \quad \uniad~\citep{uniad} & 97.8 & 91.9 & 92.9 & \textbf{100} & 78.8 & 83.4 \\
    \quad PARA-Drive~\citep{paradrive} & 97.9 & 92.4 & 93.0 & 99.8 & 79.3 & 84.0 \\
    \quad Epona~\citep{epona} & 97.9 & 95.1 & 93.8 & 99.9 & 80.4 & 86.2 \\
    \quad OneVL~\citep{onevl} & 98.5 & 96.9 & 95.8 & \textbf{100} & 82.2 & 88.4 \\
    \quad DriveLaW~\citep{drivelaw} & 99.0 & 97.1 & 96.7 & \textbf{100} & 81.3 & 89.1 \\
    \quad AutoVLA~\citep{autovla} & 98.4 & 95.6 & \textbf{98.0} & 99.9 & 81.9 & 89.1 \\
    \quad DriveVLA-W0~\citep{drivevlaw0} & 98.7 & 99.1 & 95.3 & 99.3 & 83.3 & 90.2 \\
    \quad Qwen-Drive-1.0~\citep{qwendrive} & 98.6 & 98.2 & 95.9 & \textbf{100} & 84.8 & 90.7 \\
    \quad ReCogDrive~\citep{recogdrive} & 97.9 & 97.3 & 94.9 & \textbf{100} & 87.3 & 90.8 \\
    \quad R2SE~\citep{r2se} & 99.0 & 97.9 & 96.4 & \textbf{100} & 86.8 & 91.6 \\
    \quad iPad~\citep{ipad2026} & 98.6 & 98.3 & 94.9 & \textbf{100} & 88.0 & 91.7 \\
    \quad DriveFine~\citep{drivefine} & 98.8 & 98.6 & 96.2 & \textbf{100} & 86.9 & 91.8 \\*
    \quad DrivoR-Scale~\citep{kirby2026drivor} & \textbf{99.1} & \textbf{99.2} & 96.9 & \textbf{100} & \textbf{91.6} & \textbf{94.6} \\
    \midrule
    \multicolumn{7}{@{}l}{\textit{Camera input, no human demonstrations}} \\*
    \quad DriveZero~\citep{he2026drivezero} & 99.0 & 99.2 & 96.0 & \textbf{100} & 93.1 & 94.8 \\
    \quad DriveZero-Scale~\citep{he2026drivezero} & \textbf{99.2} & 99.4 & \textbf{97.3} & 99.9 & 92.5 & 95.3 \\*
    \quad \textbf{Ours} & 98.9 & \textbf{99.5} & 92.2 & 99.9 & \textbf{96.9} & \textbf{95.5} \\
    \bottomrule
    \end{tabular}
\end{table}

\begin{table}[!htb]
\centering
\captionsetup{font=normalsize}
\scriptsize
\renewcommand{\arraystretch}{1.05}
\setlength{\tabcolsep}{2.5pt}
    \caption{Per-subscore breakdown on the NAVSIMv2 \texttt{navhard} split.
    EPDMS aggregates NC, DAC, DDC, TLC, EP, TTC, LK, HC, and EC over two
    stages; \emph{S.} is the per-stage EPDM score. Stage~2 re-scores under
    Gaussian-Splatting-based ego-state perturbations. Baseline rows are
    published numbers; \textit{italicized} rows are privileged references.
    Our rows are measured under the standard protocol.}
    \label{tab:supp-navhard-subscores}
    \begin{tabular}{@{}l c c ccccccccc c c@{}}
    \toprule
    & & & \multicolumn{9}{c}{Subscores (Stage 1 / Stage 2)} & & \\
    \cmidrule(lr){4-12}
    Method & Demo & Stage & NC & DAC & DDC & TLC & EP & TTC & LK & HC & EC & S. & EPDMS \\
    \midrule
    \multicolumn{14}{@{}l}{\textit{Privileged reference (ground-truth scene state)}} \\*
    \quad \textit{PDM-Closed~\citep{dauner2023pdm}} & \cmark & S1 & 94.4 & 98.8 & 100 & 99.5 & 100 & 93.5 & 99.3 & 87.7 & 36.0 & -- & \multirow{2}{*}{\textit{56.6}} \\*
    \quad & & S2 & 90.5 & 90.6 & 95.4 & 98.4 & 100 & 86.6 & 74.2 & 91.9 & 29.7 & -- & \\
    \quad \textit{DriveRL~\citep{he2026drivezero}} & \xmark & S1 & 99.1 & 99.6 & 100.0 & 99.6 & 91.1 & 98.7 & 99.3 & 90.9 & 18.2 & 84.1 & \multirow{2}{*}{\textit{55.6}} \\*
    \quad & & S2 & 94.4 & 92.0 & 96.4 & 98.1 & 89.1 & 89.1 & 73.8 & 88.5 & 10.9 & 65.4 & \\
    \quad \textit{Trajectory teacher (ours)} & \xmark & S1 & 99.1 & 99.8 & 100 & 99.6 & 93.0 & 99.1 & 98.4 & 97.8 & 70.2 & 92.2 & \multirow{2}{*}{\textit{62.1}} \\*
    \quad & & S2 & 92.2 & 90.4 & 93.7 & 98.1 & 90.4 & 87.0 & 63.5 & 92.9 & 54.1 & 67.5 & \\
    \quad \textit{Trajectory teacher (ours, IDM selection)} & \xmark & S1 & 100 & 100 & 100 & 99.6 & 95.3 & 99.6 & 98.9 & 97.8 & 76.9 & 94.6 & \multirow{2}{*}{\textit{65.6}} \\*
    \quad & & S2 & 93.6 & 90.0 & 92.7 & 98.4 & 92.4 & 86.6 & 67.7 & 91.4 & 54.7 & 69.5 & \\
    \midrule
    \multicolumn{14}{@{}l}{\textit{Camera input, human demonstrations}} \\*
    \quad \ltf~\citep{ltf,dauner2024navsim} & \cmark & S1 & 96.2 & 79.6 & 99.1 & 99.6 & 84.1 & 95.1 & 94.2 & 97.6 & 79.1 & -- & \multirow{2}{*}{25.1} \\*
    \quad & & S2 & 77.8 & 70.2 & 84.3 & 98.1 & 85.1 & 75.7 & 45.4 & 95.7 & 76.0 & -- & \\
    \quad NavFormer~\citep{navformer} & \cmark & S1 & 96.2 & 92.4 & 95.7 & 99.6 & 83.8 & 96.0 & 94.7 & 96.4 & 60.9 & -- & \multirow{2}{*}{34.1} \\*
    \quad & & S2 & 85.7 & 81.0 & 83.5 & 97.6 & 90.1 & 82.4 & 48.2 & 94.9 & 48.4 & -- & \\
    \quad RAP~\citep{feng2025rap} & \cmark & S1 & 97.1 & 94.4 & 98.8 & 99.8 & 83.9 & 96.9 & 94.7 & 96.4 & 66.2 & -- & \multirow{2}{*}{39.6} \\*
    \quad & & S2 & 83.2 & 83.9 & 87.4 & 98.0 & 86.9 & 80.4 & 52.3 & 95.2 & 52.4 & -- & \\
    \quad GuideFlow~\citep{guideflow} & \cmark & S1 & 99.6 & 98.0 & 99.4 & 99.3 & 79.7 & 99.3 & 94.9 & 97.1 & 58.2 & -- & \multirow{2}{*}{51.5} \\*
    \quad & & S2 & 91.4 & 89.5 & 95.2 & 98.9 & 77.5 & 89.6 & 52.6 & 93.6 & 51.0 & -- & \\
    \quad SimScale~\citep{tian2026simscale} & \cmark & S1 & 99.6 & 99.1 & 99.9 & 100.0 & 69.6 & 99.6 & 95.8 & 95.6 & 28.4 & -- & \multirow{2}{*}{53.2} \\*
    \quad & & S2 & 94.5 & 94.2 & 95.8 & 99.2 & 75.8 & 92.8 & 60.1 & 96.1 & 43.2 & -- & \\
    \quad DrivoR-Scale~\citep{kirby2026drivor} & \cmark & S1 & 99.1 & 98.2 & 99.3 & 99.8 & 75.4 & 98.7 & 94.9 & 97.6 & 70.2 & -- & \multirow{2}{*}{54.6} \\*
    \quad & & S2 & 92.3 & 91.6 & 97.3 & 99.1 & 75.7 & 90.6 & 56.1 & 98.4 & 44.7 & -- & \\
    \midrule
    \multicolumn{14}{@{}l}{\textit{Camera input, no human demonstrations}} \\*
    \quad ZTRS~\citep{ztrs} & \xmark & S1 & 98.9 & 97.6 & 100.0 & 100.0 & 66.7 & 98.9 & 96.2 & 96.7 & 44.0 & -- & \multirow{2}{*}{48.1} \\*
    \quad & & S2 & 91.1 & 90.4 & 95.8 & 99.0 & 63.6 & 89.8 & 60.4 & 97.6 & 66.1 & -- & \\
    \quad GigaPixel~\citep{rowe2026gigapixel} & \xmark & S1 & 99.4 & 95.8 & 99.4 & 99.8 & 68.1 & 99.6 & 91.8 & 97.6 & 49.8 & 77.8 & \multirow{2}{*}{50.1} \\*
    \quad & & S2 & 93.7 & 92.9 & 96.0 & 98.9 & 62.6 & 90.7 & 60.2 & 98.2 & 58.8 & 63.5 & \\
    \quad DriveZero~\citep{he2026drivezero} & \xmark & S1 & 98.7 & 98.9 & 99.9 & 99.8 & 79.5 & 98.7 & 95.8 & 97.6 & 51.6 & 84.1 & \multirow{2}{*}{51.5} \\*
    \quad & & S2 & 90.5 & 89.2 & 94.6 & 98.6 & 81.2 & 88.3 & 57.2 & 94.5 & 41.3 & 60.8 & \\
    \quad DriveZero-Scale~\citep{he2026drivezero} & \xmark & S1 & 98.7 & 99.3 & 100.0 & 99.3 & 76.5 & 98.4 & 95.3 & 97.1 & 44.4 & 82.3 & \multirow{2}{*}{57.1} \\*
    \quad & & S2 & 94.6 & 95.6 & 97.7 & 98.4 & 76.2 & 92.7 & 59.9 & 96.1 & 44.0 & 69.1 & \\
    \midrule
    \quad \textbf{Ours} & \xmark & S1 & 98.9 & 100.0 & 100.0 & 99.6 & 84.9 & 98.4 & 97.8 & 97.8 & 70.2 & 89.1 & \multirow{2}{*}{59.4} \\*
    \quad & & S2 & 94.0 & 90.2 & 93.6 & 98.0 & 85.2 & 87.2 & 62.5 & 95.4 & 53.8 & 66.5 & \\
    \bottomrule
    \end{tabular}
\end{table}